\documentclass{article}

\usepackage[final]{neurips_2021}

\usepackage[utf8]{inputenc} 
\usepackage[T1]{fontenc}    
\usepackage{hyperref}       
\usepackage{url}            
\usepackage{booktabs}       
\usepackage{amsfonts}       
\usepackage{nicefrac}       
\usepackage{microtype}      
\usepackage{xcolor}         
\usepackage{amsmath}
\usepackage{amssymb}
\usepackage{bm}
\usepackage{nicefrac}
\usepackage{enumitem}
\usepackage{microtype}
\usepackage{subfig}
\usepackage{multirow} 
\usepackage{epsfig}
\usepackage{mathtools}
\usepackage{tikz}
\usetikzlibrary{bayesnet}
\usetikzlibrary{arrows}

\hypersetup{
    colorlinks,
    linkcolor={red!90!black},
    citecolor={blue!50!black},
    urlcolor={blue!80!black}
}

\newcommand{\s}{\mathbf{s}}

\newcommand{\x}{\mathbf{x}}

\newcommand{\zbf}{\mathbf{z}}

\newcommand{\f}{\mathbf{f}}

\newcommand{\R}{\mathbb{R}}
\newcommand{\Tbb}{\mathbb{T}}

\newcommand{\Cbb}{\mathbb{C}}

\newcommand{\Mcal}{\mathcal{M}}

\newcommand{\Nbb}{\mathbb{N}}

\newtheorem{Theorem}{Theorem}

\newtheorem{Lemma}{Lemma}

\newcommand{\po}{\mathbb{P}}

\renewcommand{\geq}{\geqslant}
\renewcommand{\leq}{\leqslant}
\newcommand\eqsp{\,}
\newcommand\rmd{\mathrm{d}}

\newcommand{\ve}[1]{\mathbf{\bm{#1}}}  
\newcommand{\m}[1]{\mathbf{\bm{#1}}}  

\DeclareMathOperator{\E}{\mathbb{E}}
\DeclareMathOperator{\vect}{vec}

\newcommand{\modelname}{\texorpdfstring{$\Delta$-SNICA }{Delta-SNICA }}

\title{Disentangling Identifiable Features from Noisy Data with Structured Nonlinear ICA}

\author{%
  Hermanni H{\"a}lv{\"a}$^{1}$ \thanks{hermanni.halva@helsinki.fi}
  \qquad Sylvain Le Corff$^{2}$ 
  \qquad Luc Leh{\'e}ricy$^{3}$ 
  \AND Jonathan So$^{4}$ 
  \qquad Yongjie Zhu$^{1}$ 
  \qquad Elisabeth Gassiat$^{5}$ \thanks{Equal senior authorship}
  \qquad Aapo Hyv{\"a}rinen$^{1}$ \footnotemark[2] \\ \ \\
  $^{1}$Department of Computer Science, University of Helsinki, Finland \\
  $^{2}$ Samovar, T\'el\'ecom SudParis, d\'epartement CITI, Institut Polytechnique de Paris, Palaiseau, France \\
  $^{3}$Laboratoire J. A. Dieudonné, Université Côte d’Azur, CNRS, 06100, Nice, France \\
  $^{4}$Department of Engineering, University of Cambridge, UK \\
  $^{5}$Universit\'e Paris-Saclay, CNRS, Laboratoire de math\'ematiques d'Orsay, 91405, Orsay, France\\ 
}

\begin{document}

\maketitle

\begin{abstract}
We introduce a new general identifiable framework for principled disentanglement referred to as Structured Nonlinear Independent Component Analysis (SNICA). Our contribution is to extend the identifiability theory of deep generative models for a very broad class of structured models. While previous works have shown identifiability for specific classes of time-series models, our theorems extend this to more general temporal structures as well as to models with more complex  structures such as spatial dependencies. In particular, we establish the major result that identifiability for this framework holds even in the presence of noise of unknown distribution. Finally, as an example of our framework's flexibility, we introduce the first nonlinear ICA model for time-series that combines the following very useful properties: it accounts for both nonstationarity and autocorrelation in a fully unsupervised setting;  performs dimensionality reduction;  models hidden states; and  enables principled estimation and inference by variational maximum-likelihood.
\end{abstract}

\section{Introduction}

A central tenet of unsupervised deep learning is that noisy and high dimensional real world data is generated by a nonlinear transformation of lower dimensional latent factors. Learning such lower dimensional features is valuable as they may allow us to understand complex scientific observations in terms of much simpler, semantically meaningful, representations \citep{morioka2020nonlinear,zhou2020learning}. Access to a ground truth generative model and its latent features would also greatly enhance several other downstream tasks such as classification \citep{klindt2020towards,Banville21}, transfer learning \citep{Khemakhem20NIPS}, as well as causal inference \citep{Monti19UAI,wu2020causal}. 

A recently popular approach to deep representation learning has been to learn \textit{disentangled} features. Whilst not rigorously defined, the general methodology has been to use deep generative models such as VAEs \citep{kingma2014autoencoding, higgins2017} to estimate semantically distinct factors of variation that generate and encode the data. A substantial problem with the vast majority of work on disentanglement learning is that the models used are not \textit{identifiable} -- that is, they do not learn the true generative features, even in the limit of infinite data -- in fact, this task has been proven impossible without inductive biases on the generative model \citep{Hyva99NN,locatello19a}. Lack of identifiability plagues deep learning models broadly and has been implicated as one of the reasons for unexpectedly poor behaviour when these models are deployed in real world applications \citep{DAmour2020}. Fortunately, in many applications the data have dependency structures, such as temporal dependencies 
which introduce inductive biases. Recent advances in both identifiability theory and practical algorithms for nonlinear ICA \citep{Hyva16NIPS, Hyva17AISTATS, Halva20UAI, Morioka21AISTATS, klindt2020towards, oberhauser2021nonlinear} exploit this and offer a principled approach to disentanglement for such data. Learning statistically independent nonlinear features in such models is well-defined, i.e.\ those models are identifiable. 

However, the existing nonlinear ICA models suffer from numerous limitations. First, they only exploit specific types of temporal structures, such as either temporal dependencies or nonstationarity. Second, they often work under the assumption that some 'auxiliary' data about a \textit{latent} process is observed, such as knowledge of the switching points of a nonstationary process as in \citet{Hyva16NIPS,Khemakhem20iVAE} 
. Furthermore, all the nonlinear ICA models cited above, with the exception of \citet{Khemakhem20iVAE}, assume that the data are fully observed and noise-free, even though observation noise is very common in practice, and even \citet{Khemakhem20iVAE} assumes the noise distribution to be exactly known. This approach of modelling observation noise explicitly is in stark contrast to the approach taken in papers, such as \citet{locatello20a}, who instead consider general stochasticity of their model to be captured by latent variables -- this approach would be ill-suited to the type of denoising one would often need in practice. Lastly, the identifiability theorems in previous nonlinear ICA works usually restrict the latent components to a specific class of models such as exponential families (but see \citet{Hyva17AISTATS}).

In this paper we introduce a new framework for identifiable disentanglement, Structured Nonlinear ICA (SNICA), which removes each of the aforementioned limitations in a single unifying framework. Furthermore, the framework guarantees identifiability of a rich class of nonlinear ICA models that is able to exploit dependency structures of any arbitrary order and thus, for instance, extends to spatially structured data. This is the first major theoretical contribution of our paper. 

The second important theoretical contribution of our paper proves that models within the SNICA framework are identifiable even in the presence of additive output noise of \textit{arbitrary, unknown} distribution. We achieve this by extending the theorems by \citet{gassiat:lecorff:lehericy:2019, gassiat:lecorff:lehericy:2020}. The subsequent practical implication is that SNICA models can perform dimensionality reduction to identifiable latent components and de-noise observed data. We note that noisy-observation part of the identifiability theory is not even limited to nonlinear ICA but applies to any system observed under noise.

Third, we give mild sufficient conditions, relating to the strength and the non-Gaussian nature of the temporal or spatial dependencies,
enabling identifiability of nonlinear independent components in this general framework. An important implication is that our theorems can be used, for example, to develop models for disentangling identifiable features from spatial or spatio-temporal data.

As an example of the flexibility of the SNICA framework, we present a new nonlinear ICA model called \modelname. It achieves the following very practical properties which have previously been unattainable in the context of nonlinear ICA: the ability to account for both nonstationarity and autocorrelation in a fully unsupervised setting; ability perform dimensionality reduction; model latent states; and to enable principled estimation and inference by variational maximum-likelihood methods. We demonstrate the practical utility of the model in an application to noisy neuroimaging data that is hypothesized to contain meaningful lower dimensional latent components and complex temporal dynamics. 

\section{Background}

We start by giving some brief background on Nonlinear ICA and identifiability. Consider a model where the distribution of observed data $\ve x$  is given by $p_X(\ve x; \ve \theta)$ for some parameter vector $\ve \theta$. This model is called identifiable if the following condition is fulfilled:
\begin{align}
    \forall (\ve \theta, \ve \theta') \qquad p_X(\ve x; \ve \theta) = p_X(\ve x; \ve \theta') \Rightarrow \ve \theta = \ve \theta '\,.
\end{align}
In other words, based on the observed data distribution alone, we can \textit{uniquely} infer the parameters that generated the data. 
For models parameterized with some nonparametric function estimator $\ve f$, such as a deep neural network, we can replace $\ve \theta$ with $\ve f$ in the equation above. In practice, identifiability might hold for some parameters, not all; and parameters might be identifiable up to some more or less trivial indeterminacies, such as scaling.

In a typical nonlinear ICA setting we observe some $\ve x \in \R^N$ which has been generated by an invertible nonlinear mixing function $\ve f$ from latent independent components $\ve s \in \R^N$, with $p(\ve s)=\prod_{i=1}^N p(s^{(i)})$, as per:
\begin{align}
    \ve x = \ve f(\ve s) \label{eq:mix}\,,
\end{align}

Identifiability of $\ve f$ would then mean that we can in theory find the true $\ve f$, and subsequently the true data generating components. Unfortunately, without some additional structure this model is unidentifiable, as shown by \citet{Hyva99NN}: there is an infinite number of possible solutions and these have no trivial relation with each other. To solve this problem, previous work \citep{Sprekeler14,Hyva16NIPS, Hyva17AISTATS} developed models with temporal structure. Such time series models were generalized and expressed in a succinct way by \citet{Hyva19AISTATS,Khemakhem20iVAE} by assuming the independent components are \textit{conditionally} independent upon some observed auxiliary variable $u_t$: 
$ 
p(\ve s_t | u_t) = \prod_{i=1}^N p(s_t^{(i)}|u_t)\,.
$ 
In a time series context, the auxiliary variable might be history, e.g. $u_t=\ve x_{t-1}$, or the index of a time segment to model nonstationarity (or piece-wise stationarity). 
(It could also be data from another modality, such as audio data used to condition video data \citep{arandjelovic2017look}.)

Notice that the mixing function $\ve f$ in \eqref{eq:mix} is assumed bijective and thus \textit{identifiable} dimension reduction is not possible in most of the models discussed above. The only exceptions, we are aware of, are \citet{Khemakhem20iVAE, klindt2020towards} who choose $\ve f$ as injective rather than bijective. Further, \citet{Khemakhem20iVAE} assume additive noise on the observations 
$ 
    \ve x = \ve f(\ve \s) + \ve \varepsilon\,,
$ 
which allows to estimate posterior of $\ve s$ by an identifiable VAE (iVAE). We will take a similar strategy in what follows.

\section{Definition of Structured Nonlinear ICA}\label{sec:model}

In this section, we first present the new framework of Structured Nonlinear ICA (SNICA) -- a broad class of models for identifiable disentanglement and learning of independent components when data has structural dependencies. Next, we give an example of a particularly useful specific model that fits within our framework, called \modelname, by using switching linear dynamical latent processes.

\subsection{Structured Nonlinear ICA framework} \label{sec:snica_framework}
Consider observations $(\ve x_t)_{t \in \Tbb} = ((x_t^{(1)},\dots,x_t^{(M)}))_{t \in \Tbb}$ where $\Tbb$ is a discrete indexing set of arbitrary dimension. For discrete time-series models, like previous works, $\Tbb$ would be a subset of $\Nbb$. Crucially, however, we allow it to be any arbitrary indexing variable that describes a desired structure. For instance, $\Tbb$ could be a subset of $\Nbb^2$ for spatial data.

We assume the data is generated according the following nonlinear ICA model.
First, there exist latent
components $\ve s^{(i)} = (s_t^{(i)})_{t \in \Tbb}$ for $i \in \{1, \dots, N\}$ where for any $t, t' \in \Tbb$, the distributions of $(\ve s^{(i)}_t)_{1 \leq i \leq N}$ and $(\ve s^{(i)}_{t'})_{1 \leq i \leq N}$ are the same, which is a weak form of \textit{stationarity}. Second, we assume that for any $m \in \Nbb^*$ and $(t_1, \dots, t_m) \in \Tbb^m$,
$ 
    p(\ve s_{t_1}, \dots, \ve s_{t_m})
	= \prod_{i=1}^N p(s_{t_1}^{(i)}, \dots, s_{t_m}^{(i)})
$: 
 that is, the components are unconditionally \textit{independent}.
We further assume 
that the 
nonlinear mixing function $\f:\R^N \rightarrow \R^M$ with $M \geq N$
is injective, 
so
there may be more observed variables than components.
Finally, denote observational noise by $\ve \varepsilon_t \in \R^M$ 
and assume that they are i.i.d.\ for all $t \in \Tbb$ and 
independent of the signals $\ve s^{(i)}$. Putting these together, we assume the mixing model where for each $t \in \Tbb$,
\begin{equation}
\label{eq:mixing}
	\ve x_t = \f(\ve s_t) + \ve \varepsilon_t \eqsp,
\end{equation}
where $\ve s_t = (s^{(1)}_t, \dots, s^{(N)}_t)$. Importantly, $\ve \varepsilon_t$ can have any arbitrary unknown distribution, even with dependent entries; in fact, it may even not have finite moments.



The main appeal of this framework is that, under the conditions  given in next section, we can now guarantee identifiability for a very broad and rich class of models.

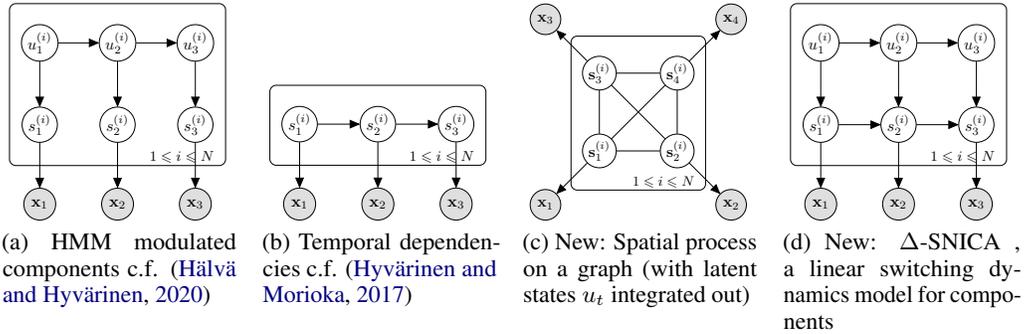
\begin{figure}
\subfloat[HMM modulated components c.f. \citep{Halva20UAI}\label{fig:hmm}]{
\resizebox{0.21\linewidth}{!}{%
	\begin{tikzpicture}
	\node[obs] (x1) {$\x_1$};%
	\node[obs, right=of x1] (x2) {$\x_2$};%
	\node[obs, right=of x2] (x3) {$\x_3$};%

	\node[latent, above=of x1, fill] (s1) {$s_1^{(i)}$}; %
	\node[latent, above=of x2, fill] (s2) {$s_2^{(i)}$}; %
	\node[latent, above=of x3, fill] (s3) {$s_3^{(i)}$}; %

	\node[latent, above=of s1, fill] (u1) {$u_1^{(i)}$}; %
	\node[latent, above=of s2, fill] (u2) {$u_2^{(i)}$}; %
	\node[latent, above=of s3, fill] (u3) {$u_3^{(i)}$}; %

	\plate [inner sep=.25cm,yshift=.2cm] {plate1} {(s1)(s2)(s3)(u1)(u2)(u3)} {$1 \leq i \leq N$}; %
 	\edge {s1}{x1}  
	\edge {s2}{x2}
	\edge {s3}{x3}
 	\edge {u1}{s1}  
	\edge {u2}{s2}
	\edge {u3}{s3}
	\edge {u1}{u2}
	\edge {u2}{u3}
	\end{tikzpicture}
	}
}
\quad
\subfloat[Temporal dependencies c.f. \citep{Hyva17AISTATS}\label{fig:pcl}]{
\resizebox{0.21\linewidth}{!}{%
	\begin{tikzpicture}
	\node[obs] (x1) {$\x_1$};%
	\node[obs, right=of x1] (x2) {$\x_2$};%
	\node[obs, right=of x2] (x3) {$\x_3$};%

	\node[latent, above=of x1, fill] (s1) {$s_1^{(i)}$}; %
	\node[latent, above=of x2, fill] (s2) {$s_2^{(i)}$}; %
	\node[latent, above=of x3, fill] (s3) {$s_3^{(i)}$}; %

	\plate [inner sep=.25cm,yshift=.2cm] {plate1} {(s1)(s2)(s3)} {$1 \leq i \leq N$}; %
 	\edge {s1}{x1}  
	\edge {s2}{x2}
	\edge {s3}{x3}
	\edge {s1}{s2}
	\edge {s2}{s3}
	\end{tikzpicture}
	}
}
\quad
\subfloat[New: Spatial process on a graph (with latent states $u_t$ integrated out) 
\label{fig:fully connected}\label{fig:fullyconnected}]{
\resizebox{0.21\linewidth}{!}{%
	\begin{tikzpicture}

	\node[latent, fill] (s1) {$\ve s_1^{(i)}$}; %
	\node[latent, right=of s1, fill] (s2) {$\ve s_2^{(i)}$}; %

	\node[latent, above=of s1, fill] (s3) {$\ve s_3^{(i)}$}; %
	\node[latent, above=of s2, fill] (s4) {$\ve s_4^{(i)}$}; %
	
	\node[obs, below left=of s1] (x1) {$\ve x_1$}; %
	\node[obs, below right=of s2] (x2) {$\ve x_2$}; %
	\node[obs, above left=of s3] (x3) {$\ve x_3$}; %
	\node[obs, above right=of s4] (x4) {$\ve x_4$}; %

	\plate [inner sep=.25cm,yshift=.2cm] {plate1} {(s1)(s2)(s3)(s4)} {$1 \leq i \leq N$}; %
 	\edge [-] {s1} {s2}
    \edge [-] {s2} {s4}
    \edge [-] {s3} {s4}  
    \edge [-] {s1} {s3}  
    \edge [-] {s2} {s3}  
    \edge [-] {s1} {s4}  

    \edge {s1} {x1}  
    \edge {s2} {x2}  
    \edge {s3} {x3}  
    \edge {s4} {x4}

	\end{tikzpicture}
	}
}
\quad
\subfloat[New: \modelname, a linear switching dynamics model for components\label{fig:slds}]{
\resizebox{0.21\linewidth}{!}{%
	\begin{tikzpicture}
	\node[obs] (x1) {$\x_1$};%
	\node[obs, right=of x1] (x2) {$\x_2$};%
	\node[obs, right=of x2] (x3) {$\x_3$};%

	\node[latent, above=of x1, fill] (s1) {$s_1^{(i)}$}; %
	\node[latent, above=of x2, fill] (s2) {$s_2^{(i)}$}; %
	\node[latent, above=of x3, fill] (s3) {$s_3^{(i)}$}; %

	\node[latent, above=of s1, fill] (u1) {$u_1^{(i)}$}; %
	\node[latent, above=of s2, fill] (u2) {$u_2^{(i)}$}; %
	\node[latent, above=of s3, fill] (u3) {$u_3^{(i)}$}; %

	\plate [inner sep=.25cm,yshift=.2cm] {plate1} {(s1)(s2)(s3)(u1)(u2)(u3)} {$1 \leq i \leq N$}; %
 	\edge {s1}{x1}  
	\edge {s2}{x2}
	\edge {s3}{x3}
 	\edge {u1}{s1}  
	\edge {u2}{s2}
	\edge {u3}{s3}
	\edge {u1}{u2}
	\edge {u2}{u3}
	\edge {s1}{s2}
	\edge {s2}{s3}
	\end{tikzpicture}
	}
}
{\caption{Graphical models for the SNICA framework}
}
\end{figure}

First, notice that all previous Nonlinear ICA time-series models can be reformulated and often improved upon when viewed through this new unifying framework. In other words, we can create models that are very much like those previous works, and capture their dependency profiles, but with the changes that by assuming unconditional independence and output noise we now allow them to perform dimension reduction (this does also require some additional assumptions needed in our identifiability theorems below). To see this, consider the model in \citet{Halva20UAI} which captures nonstationarity in the independent components through a global hidden Markov chain. We can transform this model into the SNICA framework if we instead model \textit{each} independent component as its own HMM (Figure \ref{fig:hmm}), with the added benefit that we now have marginally independent components and are able to perform dimensionality reduction into low dimensional latent components. Nonlinear ICA with time-dependencies, such as in an autoregressive model, proposed by \citet{Hyva17AISTATS} is also a special case of our framework (Figure \ref{fig:pcl}), but again with the extension of dimensionality reduction. Furthermore, this framework allows for a plethora of new Nonlinear ICA models to be developed. As described above, these do not have to be limited to time-series but could for instance be a process on a two-dimensional graph with appropriate (in)dependencies (see Figure \ref{fig:fullyconnected}). However, we now proceed to introduce a particularly useful time-series model using our framework.



\subsection{\modelname: Nonlinear ICA with switching linear dynamical systems} \label{sec:sld-snica}

While the above framework has great generality, any practical application will need a specific model. Next we propose one which combines the following properties of previous nonlinear ICA models into a single model: ability to account for both nonstationarity and autocorrelation in a fully unsupervised setting, to perform dimensionality reduction and model hidden states. Real world processes, such as video/audio data, financial time-series, and brain signals, exhibit these properties -- disentangling latent features in such data would hence be very useful. 

Our new model is depicted in Figure \ref{fig:slds}. The independent components are generated by a Switching Linear Dynamical System (SLDS) \citep{Ackerson1968SLDS, ChangSLDS, HamiltonSLDS, ghahramani2000variational} with additional latent variables to express rich dynamics. Formally, for each independent component $i \in \{1,\dots,N\}$, consider the following SLDS over some latent vector $\ve y_t^{(i)}$: 
\begin{align}
	\ve y_t^{(i)} = \ve B_{u_t}^{(i)} \ve y_{t-1}^{(i)} + \ve b_{u_t}^{(i)} + \ve \varepsilon_{u_t}^{(i)} \eqsp, \label{eq:sld}
\end{align}
where $u_t:=u_t^{(i)}$ is a state of a first-order hidden Markov chain $(u_t^{(i)})_{t=1:T}$. Crucially, we assume that the independent components at each time-point are the first elements $y_{t,1}^{(i)}$ of $\ve y_t^{(i)} = (y_{t,1}^{(i)}, \dots, y_{t,d}^{(i)})^T$, i.e. $s_t^{(i)} = y_{t,1}^{(i)}$. The rest of the elements in  $\ve y_{t}^{(i)}$ are latent variables modelling hidden dynamics.
The great utility of using such a higher-dimensional latent variable is that this model allows us, for example, as a special case, to consider higher-order ARMA processes, thus modelling each $s_t^{(i)}$ as switching between ARMA processes of an order determined by the dimensionality of $\ve y_t$.
We call the ensuing model \modelname ("Delta-SNICA", with delta as in "dynamic").





\section{Identifiability} \label{sec:identif}
In this section, we present two very general identifiability theorems for SNICA. We basically decouple the problem into two parts. First, we consider identifying the noise-free distribution of $\ve f(\ve s_t)$ from noisy data. Theorem~\ref{theo1} states conditions---on tail behaviour, non-degeneracy, and non-Gaussianity---under which it is possible to recover the distribution of a process based on noisy data with unknown noise distribution. Second, we consider demixing of the nonlinearly mixed data. Theorem~\ref{theo2} provides general conditions---on temporal or spatial dependencies, and non-Gaussianity---that allow recovery of the mixing function $\ve f$ when there is no more noise.
We then consider application of these theorems to SNICA.


\subsection{Identifiability with unknown noise distribution}
\label{subsecTheo1}

Consider the model 
\begin{equation}
\label{eq:deconvol}
	\ve x_t = \ve z_t +  \ve \varepsilon_t \eqsp,
\end{equation}
where $(\ve z_t)_{t \in \Tbb}$ is a family of random variables in $\mathbb{R}^M$ such that all $\ve z_t$, $t \in \Tbb$, have the same marginal distribution, and $(\ve \varepsilon_t)_{t \in \Tbb}$ is a family of independent (over $t$) and identically distributed random variables, independent of $(\ve z_t)_{t \in \Tbb}$. Let $P$ be the common distribution of each $\ve \varepsilon_{t}$, for $t\in \Tbb$. 
Let $t_1$ and $t_2$ in $\Tbb$, and consider the following assumptions.
\begin{itemize}
\item (A1) [Tail behaviour]
For some $\rho <3$, there exist $A$ and $B$ such that for all $\lambda\in\R^N$, 
\begin{equation*}
\mathbb{E}[\exp(\langle \lambda, \ve z_{t_1}\rangle)] \leq A \exp (B\|\lambda\|^{\rho})\eqsp.
\end{equation*}
\item (A2) [Non-degeneracy]
For any $\eta \in \Cbb^{M}$, $\mathbb{E}[\exp \{\langle \eta,\ve z_{t_2}\rangle \} \vert \;\ve z_{t_1}]$
is not the null random variable.
\item (A3) [Non-Gaussianity]
The following assertion is false: there exist a vector $\eta \in \R^M$ and independent random variables $\tilde{z}$ and $u$, such that $u$ is a non dirac Gaussian random variable and $\langle \eta, \ve z_{t_1} \rangle$ has the same distribution as $\tilde{z}+u$.
\end{itemize}
We defer the detailed discussion on the practical meaning of the assumptions (A1-A3) in the context of SNICA to Section~\ref{subsec:appliSNICA}. We next present Theorem~\ref{theo1} which establishes identifiability under unknown noise (its proof is postponed to Section~\ref{sec:proof:identwithnoise} in the Supplementary Material):
\begin{Theorem}
\label{theo1}
Assume that assumptions (A1), (A2) and (A3) hold for some $(t_1,t_2)\in\Tbb^2$.
Then, up to translation, for all $m\geq 2$, for all $(t_3, \dots, t_m) \in \Tbb^{m-2}$, 
the application that associates the distribution of $(\ve z_{t_1},\ldots,\ve z_{t_m})$
and $P$ to the distribution of $(\ve x_{t_1},\ldots,\ve x_{t_m})$ is one-to-one. 
\end{Theorem}
Here, up to translation means that adding a constant vector to all $\varepsilon_t$,  and substracting this constant to all $\ve z_t$, $t\in\{t_1,\ldots, t_m\}$, does not change the distribution of $(\ve x_{t_1},\ldots,\ve x_{t_m})$. The proof of Theorem~\ref{theo1} extends that of Theorem~1 in \citep{gassiat:lecorff:lehericy:2019}, see also \citep{gassiat:lecorff:lehericy:2020}, which assumed sub-Gaussian noise-free data. Our extension allows the noise-free data to have heavier tails, which is important since (noise-free) data in many real-world applications is super-Gaussian, i.e.\ heavy-tailed, as is well-known in work on linear ICA \citep{Hyvabook}.  


Importantly, there is no assumption on the unknown noise distribution in Theorem~\ref{theo1}. In fact, it does not even assume a mixing as in ICA, and thus extends greatly outside of the framework of this paper.


\subsection{Identifiability of the mixing function}
\label{subsec:Theo2}
Based on Theorem~\ref{theo1}, it is possible to recover the distribution of the noise-free data in SNICA in \eqref{eq:mixing} by setting $\ve z_t=\ve f(\ve s_t)$. Next, we consider under which conditions the mixing function $\ve f$ is identifiable.
Denote by $S = S^{(1)} \times \dots \times S^{(N)}$ the support of the distribution of  all $\ve s_{t}$. 
We consider the situation where each $S^{(i)} \subset \R$, $1 \leq i \leq N$, is connected, so that each $S^{(i)}$ is an interval. 
We assume moreover that the injective mixing function $\ve f$ is a ${\cal C}^2$ diffeomorphism between $S$ and a ${\cal C}^2$ differentiable manifold ${\cal M} \subset \R^M$.
Formally, this means that there exists an atlas $\{\varphi_\vartheta : U_\vartheta \rightarrow \R^N \}_{\vartheta \in \Theta}$ of ${\cal M} $ such that for all $\vartheta, \vartheta' \in \Theta$, the map $\varphi_{\vartheta} \circ \varphi_{\vartheta'}^{-1}$ is a ${\cal C}^2$ map, and $\ve f$ is a bijection $\R^N \rightarrow \Mcal$ such that for all $\vartheta \in \Theta$, $\varphi_\vartheta \circ \ve f$ and $\ve f^{-1} \circ \varphi_\vartheta^{-1}$ have continuous second derivatives. The sets $U_\vartheta$, $\vartheta \in \Theta$, cover $\Mcal$ and are open in $\Mcal$. The proof of Theorem~\ref{theo2} is postponed to Section~\ref{sec:proof:f:manifold} in the Supplementary Material.
\begin{Theorem}
\label{theo2}
Assume that there exist $m \geq 2$ and $(t_1, \dots, t_m) \in \Tbb^m$ such that the vector
$(s_{t_1}^{(i)}, \dots, s_{t_m}^{(i)})$ has a density $p_m^{(i)}$ which is ${\cal C}^2$ on $(S^{(i)})^m$. Assume moreover that there exist
$(k,l) \in \{1, \dots, m\}^2$ with $k \neq l$ such that the following assumptions hold with
 $Q_m^{(i)} = \log p_m^{(i)}$.
\begin{itemize}
\item (B1) (Uniform $(k,l)$-dependency). For all $i \in \{1,\dots,N\}$, the set of zeros of
$ 
\frac{\partial^2}{\partial s_{t_k}^{(i)} \partial s_{t_l}^{(i)}} Q_m^{(i)}
$ 
is a meagre subset of $(S^{(i)})^m$, i.e. it contains no open subset.

\item (B2) (Local $(k,l)$-non quasi Gaussianity).
For any open subset $A \subset S^m$, there exists at most one $i \in \{1, \dots, N\}$ such that there exists a function $\alpha:\R^{m-1} \rightarrow \R$ and a constant $c \in \R$ such that for all $s\in A$,
\begin{equation} \label{eq:B2}
\frac{\partial^2}{\partial s_{t_k}^{(i)} \partial s_{t_l}^{(i)}} Q_m^{(i)}
	= c \, \alpha (s_{t_k}^{(i)}, \ve s_{(-t_k,-t_l)}^{(i)}) \alpha (s_{t_l}^{(i)}, \ve s_{(-t_k,-t_l)}^{(i)})\eqsp,
\end{equation}
where $\ve s_{(-t_k,-t_l)}^{(i)}$ is $(s_{t_1}^{(i)},\ldots,s_{t_m}^{(i)})$ without the coordinates $t_k$ and $t_l$.
\end{itemize}
Then, $\ve f^{-1}$ can be recovered up to permutation and coordinate-wise transformations from the distribution of $(\ve f(\ve s_{t_1}), \dots, \ve f(\ve s_{t_m}))$.
\end{Theorem}
%

\subsection{Applications to SNICA}
\label{subsec:appliSNICA}

In this section, we provide additional comments on the assumptions (A1-A3) and (B1-B2) and their verification in the context of SNICA.

\paragraph{Assumption (A1)} is a condition on the tails of the noise-free data: it allows tails that are somewhat heavier than Gaussian tails. It is in fact equivalent to assuming that for some $\tilde{\rho} > 3/2$, there exists $A', B' > 0$ such that for all $t > 0$, $\po(\|\ve z_{t_1}\| \geq t) \leq A' \exp(-B' t^{\tilde{\rho}})$. 

\paragraph{Assumption (A2)} is a non-degeneracy condition likely to be fulfilled for any randomly chosen SNICA model parameters. As an example,  consider a model such as Fig.~\ref{fig:fullyconnected}, where there exist hidden variables $(u_t)_{t \in \Tbb}$ taking values in a finite set $\{1, \dots, K\}$ such that the pairs of variables $(\ve z_t, u_t)$ have the same distribution for all $t \in \Tbb$, and such that conditioned on $(u_t)_{t \in \Tbb}$, the variables $(\ve z_t)_{t \in \Tbb}$ are independent and the distribution of $\ve z_t$ only depends on $u_t$. (As a special case, this model includes the temporal HMM setting described in  Fig.~\ref{fig:hmm}.) 
Let $(t_1, t_2) \in \Tbb^2$.
For all $u, v \in \{1, \dots, K\}$, let $\pi(u) = p_{u_{t_1}}(u)$ be the mass function of $u_{t_1}$, $Q(u, v) = p_{u_{t_2} | u_{t_1}}(v | u)$ be the transition matrix from $u_{t_1}$ to $u_{t_2}$, and $\gamma_{u}(\ve z) = p_{\ve z_{t_1} | u_{t_1}}(\ve z | u)$ be the density of $\ve z_{t_1}$ conditionally to $u_{t_1} = u$. By assumption, it is also the density of $\ve z_{t_2}$ conditionally to $u_{t_2} = u$. Theorem~\ref{th:A2} provides sufficient conditions for assumption (A2) to hold:
\begin{Theorem}
\label{th:A2}
Assume that $Q$ has full rank, $\min_u \pi(u) > 0$ and the $(\gamma_u)_{1 \leq u \leq K}$ are linearly independent, then (A2) is satisfied as soon as the functions $(\eta \mapsto \int \exp(\langle \eta, \ve z\rangle)\gamma_{v}(\ve z)d\ve z)_{1 \leq v \leq K}$ do not have simultaneous zeros.
\end{Theorem}
Besides the non-simultaneous zeros assumption, the assumptions of Theorem~\ref{th:A2} are reminiscent of those  used for the identifiability of non-parametric hidden Markov models, see for instance~\cite{MR3439359, lehericy2019order}.
The key element is that $\ve z_{t_1}$ and $\ve z_{t_2}$ are not independent. Thus, we see that (A2) holds if the $\pi$ and the $\gamma$ are not degenerate (in the precise sense given by Theorem~\ref{th:A2}), for the latent state models in Figs.~\ref{fig:hmm},\ref{fig:fullyconnected}.
Another situation where (A2) holds is when $\ve z_{t_2}$ is a complete statistic \citep{lehmann2006theory} in the statistical model $\{\po_{\ve z_{t_2} | \ve z_{t_1}}(\cdot | \ve z_{t_1})\}_{\ve z_{t_1}}$, where $\po_{\ve z_{t_2} | \ve z_{t_1}}(\cdot | \ve z_{t_1})$ is the distribution of $\ve z_{t_2}$ conditionally to $\ve z_{t_1}$. 
Consider the two following examples where this holds: 1) When the model $\{\po_{\ve z_{t_2} | \ve z_{t_1}}(\cdot | \ve z_{t_1})\}_{\ve z_{t_1}}$ is an exponential family. In this situation, complete statistics are known. 2) Autoregressive models with additive innovation of the form $\ve z_{t_2} = \ve h(\ve z_{t_1}) + \ve v_{t_2}$ for some bijective function $\ve h$ when the additive noise $\ve v_{t_2}$ is a complete statistic in the statistical model $\{\po_{\ve v_{t_2} | \ve z_{t_1}}(\cdot | \ve z_{t_1})\}_{\ve z_{t_1}}$ (note that $\ve v_{t_2}$ cannot be independent of $\ve z_{t_1}$ here). The case in Fig.~\ref{fig:pcl} is typically covered by this example. 

\paragraph{Assumption (A3)} states that no direction of the noise free data has a non Dirac Gaussian variable component.
It holds as soon as $\ve z_t= \ve f (\ve s_t)$ and the range of $\ve f$ 
is such that its orthogonal projection on any line is not the full line. This assumption holds for instance in the following cases: 1) The range of $\ve f$ is compact, or 2) the range of $\ve f$ is contained in a half-cylinder, that is, there exists a hyperplane such that the range of $\ve f$ is only on one side of this hyperplane and the projection of the range of $\ve f$ on this hyperplane is bounded.

\paragraph{Assumption (B1) and Assumption (B2)} are similar to those in \citep{Hyva17AISTATS,oberhauser2021nonlinear}  in the special case of time-series, i.e.\ $\Tbb = \Nbb$. (B1) then entails that there must be sufficiently strong statistical dependence between nearby time points. (B2) is a condition which excludes Gaussian processes and processes which can be trivially transformed to be Gaussian. (For treatment of the Gaussian case, see Appendix~\ref{gaussiandiscussion} in Supplementary Material.) We can further provide a simple and equivalent formulation when the independent components $\ve s^{(i)}$ follow independent and stationary HMMs with two hidden states, which is a special case of SNICA.
Denote by $\gamma^{(i)}_0$ and $\gamma^{(i)}_1$ the densities of $s^{(i)}_t$ conditionally to $\{u^{(i)}_t = 0\}$ and $\{u^{(i)}_t = 1\}$ respectively. 
\begin{Theorem}\label{th:simplehmm}
Assume that the stationary distribution $\pi$ of the hidden chain is such that $0 < \pi(0) < 1$ and that its transition matrix is invertible. Then
(B1) and (B2) are satisfied with $m=2$ if and only if on any open interval, $\gamma^{(i)}_0$ and $\gamma^{(i)}_1$ are not proportional.
\end{Theorem}
Thus, a very simple HMM leads to these conditions being verified.  \citet{Hyva17AISTATS} already showed that the conditions (B1) and (B2) also hold in the case of non-Gaussian autoregressive models. Thus, we see that our identifiability theory applies both in the case HMM's (Fig~\ref{fig:hmm}) and autoregressive models (Fig~\ref{fig:pcl}), the two principal kinds of temporal structure proposed in previous work, while extending them to further cases and combinations such as in Fig~\ref{fig:fullyconnected},\ref{fig:slds}.

\paragraph{A simplification of (B1,B2)} It is also possible to combine the assumptions  (B1) and (B2) in one, while slightly weakening the generality. The key is to notice that (\ref{eq:B2}) in (B2) implies the derivative in (B1) is zero, by setting $c=0$. But there is still the difference that (B2) considers all but one index while (B1) considers all indices $i$. If we simply assume (\ref{eq:B2}) does not hold for any $i$, we can replace (B1) and (B2) by the new condition:
\begin{itemize}
\item (B') For any open subset $A \subset S^m$ and for any $i \in \{1,\dots,N\}$, a function $\alpha:\R^{m-1} \rightarrow \R$ and a constant $c \in \R$ do not exist such that (\ref{eq:B2}) would hold for all $s\in A$.
\end{itemize}
Note that \citet{Hyva17AISTATS} defined uniform dependency and (non-)quasi-Gaussianity as two separate properties, but in fact their assumption of non-quasi-Gaussianity was weaker than ours: it did not consider all open subsets separately, which is why this simplification was not possible for them. We believe their definition of non-quasi-Gaussianity was in fact not quite sufficient to prove their theorem, and our stronger version may be needed, in line with \citet{oberhauser2021nonlinear}.

\section{Experiments} \label{sec:xperiment}

\paragraph{Estimation method} 
One challenge is that it is not practically possible to learn \modelname by exact maximum-likelihood methods. Instead, we perform learning and inference using Structured VAEs \citep{johnson2017composing} -- the current state-of-art in variational inference for structured models. Specifically, this consists of assuming that the latent posterior factorizes as per $q(\ve y_{1:T}^{(1:N)}, u_{1:T}^{(1:N)}) = \prod_{i=1}^N q(\ve y_{1:T}^{(i)}) q(u_{1:T}^{(i)})$, which allows us to optimize the resulting evidence lower bound (ELBO):
\begin{align}
    \log \widehat{\mathcal{L}}
    &= \E_{q}\bigg[\sum_{t=1}^T\log p(\ve x_t\mid\ve s_t^{(1)}, ..., \ve s_t^{(N)})\bigg] + \sum_{i=1}^N\Bigg(- \mathrm{KL}\bigg[q(u_{1:T}^{(i)})\bigg| p(u_{1:T}^{(i)})\bigg] + \mathrm{H}\bigg[q(\ve s_{1:T}^{(i)})\bigg] \nonumber\\
    &\qquad+ \E_{q}\bigg[\log p(\ve s_1^{(i)}\mid u_1^{(i)}) \bigg] + \sum_{t=2}^T \E_{q}\bigg[\log p(\ve s_t^{(i)}\mid\ve s_{t-1}^{(i)}, u_t^{(i)}) \bigg]\Bigg).
\end{align}
Since all the distributions are in conjugate exponential families (encoder neural network is used to approximate the natural parameters of the nonlinear likelihood term) efficient message passing can be used for inference, and the mixing function is learned as decoder neural network. Even though this method lacks consistency guarantees (but see \citet{Wang_2018}), we find that our model performs very well. A more detailed treatment of estimation and inference of \modelname is given in Appendix \ref{sec:apdx:estimation}. Our code will be openly available at \href{https://github.com/HHalva/snica}{https://github.com/HHalva/snica}.

\subsection{Experiments on simulated data}
The identifiability theorems stated above hold in the limit of infinite data. Additionally, a consistent estimator would be required to learn the ground-truth components. In the real world, we are limited by data and estimation methods and hence it is unclear as to what extent we are actually able to estimate identifiable components -- and whether identifiability reflects in better performance in real world tasks. To explore this, we first performed experiments on simulated data. 
 We compared the performance of our model to the current state-of-the-art, IIA-HMM \citep{Morioka21AISTATS}, as well as identifiable VAE (iVAE) \citep{Khemakhem20iVAE} and standard linear Gaussian state-space model (LGSSM). Since iVAE is not able to handle latent auxiliary variables, we allow it to "cheat" by giving it access to the true data generating latent-state, thereby creating a presumably challenging baseline~(denoted iVAE$^*$ in our figures). LGSSM was included as a naive baseline which is only able to estimate linear mixing function. 
 
\begin{figure}
    \centering
    \includegraphics[width=0.7\columnwidth]{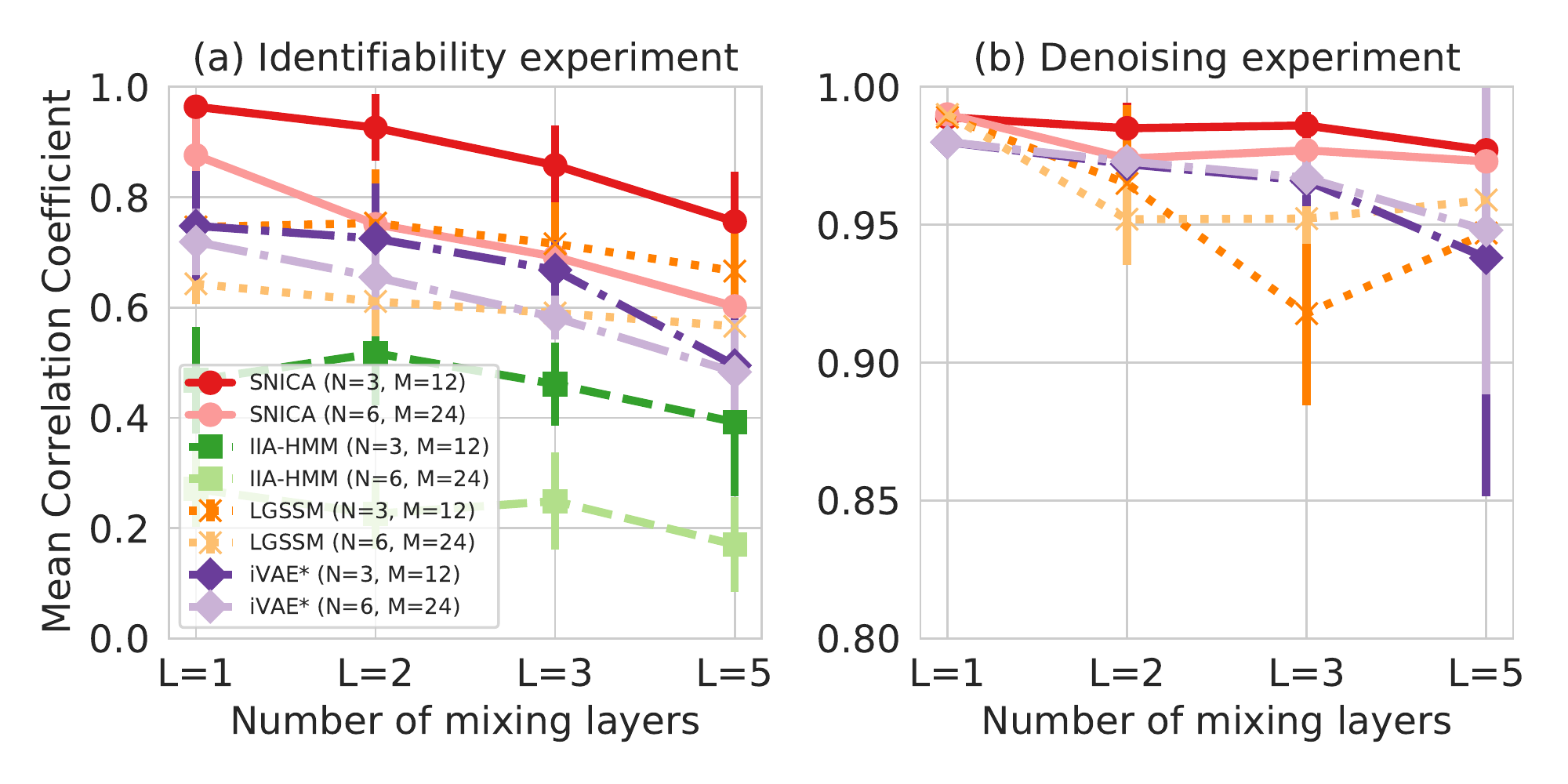}
    \caption{(a) Mean absolute correlation coefficients between ground-truth independent components and their estimates by \modelname, IIA-HMM, LGSSM and iVAE$^*$, with different orders of complexity (number of layers) and two different dimensions of observed (12, 24) and latent (6, 12) data. (b) Mean absolute correlation coefficient between estimated noise free data and ground-truth noise free data for same set of models except IIA-HMM. Please note the difference in y-axis scales.}
    \label{fig:sim}
\end{figure}

\paragraph{Investigating identifiability and consistency}
We simulated 100K long time-sequences from the \modelname\ model and computed the mean absolute correlation coefficient (MCC) between the estimated latent components and ground truth independent components (see Supplementary material for further implementation details). More precisely, to illustrate the dimensionality reduction capabilities we considered two settings where the observed data dimension $M$, was either 12 or 24 and the number of independent components, $N$ was 3 and 6, respectively. Since IIA-HMM is unable to do dimensionality reduction, we used PCA to get the data dimension to match that of the latent states. We considered four levels of mixing of increasing complexity by randomly initialized MLPs of the following number of layers: 1 (linear ICA), 2, 3, and 5. The results in Figure \ref{fig:sim}a) illustrate the clearly superior performance of our model. The especially poor performance of IIA-HMM maybe explained by lack of noise model, much simpler latent dynamics, and lost information due to PCA pre-processing. See Appendix~\ref{sec:apdx:simudetails} for further discussion and training details. 

\paragraph{Application to denoising}
\modelname is able to denoise time-series signals by learning the generative model and then performing inference on latent variables. Specifically, SVAE learns the encoder network which is used to perform inference on the posterior of the independent components. We illustrate this using the same settings as above, with the exception that we now use our learned encoder and inference to get the posterior means of the independent components and input these in to the estimated decoder to get predicted noise-free observations, denoted as $\widehat{\ve f}(\ve s_t)$ -- we measured the correlation between $\widehat{\ve f}(\ve s_t)$ and the ground-truth $\ve f(\ve s_t)$. Note that IIA-HMM, is not able to perform this task. The results in Figure \ref{fig:sim}b) show that the other models, designed to handle denoising, perform well at this task, as would be expected -- identifiability of the latent state is not necessary for good denoising performance. For LGSSM, denoising is done with the Kalman Smoother algorithm.

\subsection{Experiments on real MEG data}
To demonstrate real-data applicability, \modelname was applied to multivariate time series of electrical activity in the human brain, measured by magnetoencephalography (MEG). Recently, many studies have demonstrated the existence of fast transient networks measured by MEG in the resting state and the dynamic switching between different brain networks \citep{baker2014fast, vidaurre2017brain}. Additionally, such MEG data is high-dimensional and very noisy. Thus this data provides an excellent target for \modelname to disentangle the underlying low-dimensional components.

\begin{figure}
    \centering
    \includegraphics[width=0.7\columnwidth]{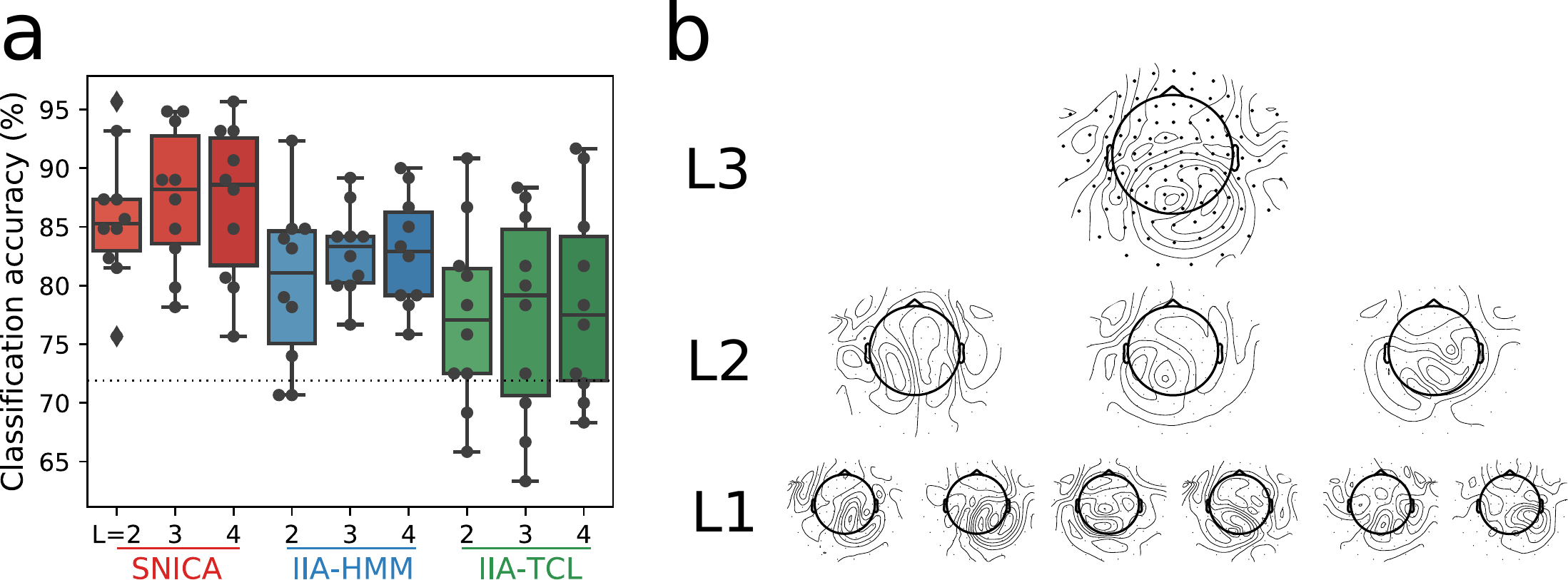}
    \caption{\modelname on MEG data. (a) Classification accuracies of linear SVMs trained with auditory-visual data to predict stimulus category, with feature extractors trained by \modelname in advance with resting-state data. 
    Each point represents a testing accuracy on a target subject (chance level: 50\%). Horizontal dotted line is PCA-only baseline. (b) Example of spatial patterns of the components learned by \modelname (L=3). Each topography corresponds to one spatial pattern. L3: approximate total spatial pattern of one third-layer unit. L2: the patterns of the three second-layer units maximally contributing to this L3 unit. L1: for each L2 unit, the two most strongly contributing first-layer units.}
    \label{fig:meg}
\end{figure}

\paragraph{Data and Preprocessing}
 We considered a resting state MEG sessions from the Cam-CAN dataset. 
 During the resting state recording, subjects sat still with their eyes closed. In the task-session data, the subjects carried out a (passive) audio–visual task including visual stimuli and auditory stimuli. We exclusively used the resting-session data for the training of the network, and task-session data was only used in the evaluation. The modality of the sensory stimulation provided a class label that we used in the evaluation, giving in total two classes. We band-pass filtered the data between 4 Hz and 30 Hz (see Supplementary Material for the details of data and settings).
\paragraph{Methods}
The resting-state data from all subjects were temporally concatenated and used for training. The number of layers of the decoder and encoder were equal and took values 2, 3, 4. We fixed the number of independent components to 5 so that our result can be fairly compared to those in \citet{Morioka21AISTATS}. 
To evaluate the obtained features, we performed classification of the sensory stimulation categories by applying feature extractors trained with (unlabeled) resting-state data to (labeled) task-session data. Classification was performed using a linear support vector machine (SVM) classifier trained on the stimulation modality labels and sliding-window-averaged features obtained for each trial. The performance was evaluated by the generalizability of a classifier across subjects. i.e., one-subject-out cross-validation. For comparison, we evaluated the baseline methods: IIA-HMM and IIA-TCL \citep{Morioka21AISTATS}. We also visualized the spatial activity patterns obtained by \modelname, using the weight vectors from encoder neural network across each layer.
\paragraph{Results}
Figure \ref{fig:meg} a) shows the classification accuracies of the stimulus categories, across different methods and the number of layers for each model. The performances by \modelname were consistently higher than those by the other (baseline) methods, which indicates the importance of the modeling of the MEG signals by \modelname. Figure \ref{fig:meg} b) shows an example of spatial patterns from the encoder network learned by the \modelname. We used the visualization method presented in \citep{Hyva16NIPS}. 
We manually picked one out of the hidden nodes from the third layer in encoder network, and plotted its weighted-averaged sensor signals, We also visualized the most strongly contributing second- and first-layer nodes. We see progressive pooling of L1 units to form left lateral frontal, right lateral frontal and parietal patterns in L2 which are then all pooled together in L3 resulting in a lateral frontoparietal pattern. Most of the spatial patterns in the third layer (not shown) are actually similar to those previously reported using MEG \citep{brookes2011investigating}. Appendix \ref{detailmeg} provides more detail to the interpretation of the \modelname results.

\section{Related work}\label{sec:relatedwork}

Previous works on nonlinear ICA have exploited autocorrelations \citep{Hyva17AISTATS, oberhauser2021nonlinear} and nonstationarities \citep{Hyva16NIPS, Halva20UAI} for identifiability. The SNICA setting provides a unifying framework which allows for both types of temporal dependencies, and further, extends identifiability to other temporal structures as well as any arbitrary higher order data structures which has not previously been considered in the context of nonlinear ICA. Another major theoretical contribution here is to show that identifiability with noise of unknown, arbitrary distribution, while previous work on noisy nonlinear ICA assumed noise of known distribution and known variance \citep{Khemakhem20iVAE}. 

Importantly, the SNICA framework is fully probabilistic and thus accommodates higher order latent variables, leading to "purely unsupervised" learning. This is in large contrast to previous research which have been developed for the case where we are able to observe some additional auxiliary variable, such as audio signals accompanying video \citep{Hyva19AISTATS,Khemakhem20iVAE,Khemakhem20NIPS}, or heuristically define the auxiliary variable based on time structure \citep{Hyva16NIPS}. 
In practice this means that we are able to estimate our models using (variational) MLE, which is more principled than the heuristic self-supervised methods in most earlier papers. The only existing frameworks allowing MLE \citep{Halva20UAI,Khemakhem20iVAE} used model restricted to exponential families, and had either no HMM or a very simple one.


The switching linear dynamical model, \modelname in Section~\ref{sec:sld-snica}, shows the above benefits in the form of a single model. That is, unlike previous nonlinear ICA models, it combines: 1) temporal dependencies and "non-stationarity" (or HMM) in a single model 2) dimensionality reduction within a rigorous maximum likelihood learning and inference framework, and 3) a separate observation equation with general observational noise. This results in a very rich, realistic, and principled model for time series.

Very recently, \citet{Morioka21AISTATS} proposed a related model by considering innovations of time series to be nonstationary. However, their model is noise-free, restricted to exponential families of at least order two, and not applicable to the spatial case, thus making our identifiability results significantly stronger. From a more practical viewpoint, their model suffers from the fact that it either does not allow for dimensionality reduction (if an HMM is used) or requires a manual segmentation (if HMM is not used). Nor does it have a clear distinction into a state dynamics equation and a measurement equation which allows for cleaning or denoising of the data. 



\paragraph{Limitations} \label{sec:limits}
Our identifiability theory makes some restrictive assumptions, and it remains to be seen if they could be lifted in future work. In particular, the data is not allowed to have too heavy tails; the noise must be additive, and independent of the signal; and the practical interpretation of some of the assumptions, such as (A3) is difficult.
It is also difficult to say whether our assumption of unconditionally independent components is realistic in practice. Regarding practical applications, our specific model only scratches the surface of what is possible in this framework. In particular, we did not develop a model with spatial distributions, nor did we model non-Gaussian observational noise -- our main aim was to lay the foundations for the relevant identification theory. Future work should aim to make the estimation more efficient computationally; this is a ubiquitous problem in deep learning, but specific solutions for this concrete problem may be achievable \citep{Gresele20}. 



\section{Conclusion}
We proposed a new general framework for identifiable disentanglement, based on nonlinear ICA with very general temporal dynamics or spatial structure. Observational noise of arbitrary unknown distribution is further included. We prove identifiability of the models in this framework with high generality and mathematical rigour. For real data analysis, we propose a special case which subsumes the properties of all existing time series models in nonlinear ICA, while generalizing them in many ways (see Section~\ref{sec:relatedwork} for details). We hope this work will contribute to wide-spread application of identifiable methods for disentanglement in a highly principled, probabilistic framework.

\newpage

\acksection
The authors would like to thank Richard Turner for insightful comments and discussion on this work. The authors also wish to thank the Finnish Grid and Cloud Infrastructure (FGCI) for supporting this project with computational and data storage resources. A.H. was supported by a Fellowship from CIFAR, and the Academy of Finland. E.G. would like
to acknowledge support for this project from Institut Universitaire de France. J.S. is supported by the University of Cambridge Harding Distinguished Postgraduate Scholars Programme.

\bibliography{bib}
\bibliographystyle{apalike}

\newpage
\appendix

\section{Appendix}

\subsection{Proof of Theorem~\ref{theo1}}
\label{sec:proof:identwithnoise}
Let $m \geq 2$ and $(t_1, \dots, t_m) \in \Tbb^m$. Let $R_m$ and $\tilde{R}_m$ be two possible distributions for $(\ve z_{t_1}, \dots, \ve z_{t_m})$ that satisfy assumptions (A1), (A2) and (A3) and let $P$ and $\tilde{P}$ be two possible distributions for $\ve \varepsilon_{t_1}$. Assume that the distribution of $(\ve x_{t_1}, \dots, \ve x_{t_m})$ in the model~\eqref{eq:deconvol} is the same under $(R_m, P)$ and $(\tilde{R}_m, \tilde{P})$.

Write $\Phi_{R_m}$ the characteristic function of $R_m$, and likewise $\Phi_{\tilde{R}_m}$, $\Phi_P$ and $\Phi_{\tilde{P}}$.
Following the proof of Theorem 1 of \cite{gassiat:lecorff:lehericy:2019} on the distribution of $(\ve z_{t_1}, \ve z_{t_2})$, as in Assumption (A1) we have $\rho<3$ , by Hadamard's factorization theorem
, there exist a polynomial function $Q$ with total degree at most $2$ and a neighborhood $V$ of $0$ in $\R^M$ such that for all $\ve u \in V$,
\begin{equation}
\label{eq_noise}
\Phi_{P}(\ve u) \exp\{Q(\ve u)\}
    = \Phi_{\tilde{P}}(\ve u)\eqsp.
\end{equation}
For completeness we provide at the end of this section the sketch of proof of~\eqref{eq_noise}.

Writing the characteristic function of $(\ve z_{t_1}, \dots, \ve z_{t_m})$ under the two sets of parameters yields, for all $(\ve u_1, \dots, \ve u_m) \in V^m$,
\begin{equation}
\label{eq_fn_carac_z}
\Phi_{R_m}(\ve u_1, \dots, \ve u_m) \prod_{k=1}^m \Phi_P(\ve u_k) 
    = \Phi_{\tilde{R}_m}(\ve u_1, \dots, \ve u_m) \left(\prod_{k=1}^m \Phi_P(\ve u_k)\right) \left(\prod_{k=1}^m \exp(Q(\ve u_k))\right)\eqsp.
\end{equation}
Since $\Phi_P$ is continuous and non-zero at 0, we may divide both sides by $\prod_{k=1}^m \Phi_P(\ve u_k)$ on a neighborhood of zero.
Under assumption (A1), $\Phi_{R_m}$ and $\Phi_{\tilde{R}_m}$ can be extended into multivariate analytic functions:
\begin{eqnarray*}
\Phi_{R_m}: \hspace{1cm} (\Cbb^{M})^m &\longrightarrow& \Cbb \\
(\ve u_1, \dots, \ve u_m) &\longmapsto& \int \exp \left(i \ve u_1^\top \ve z_{t_1} + \dots + i \ve u_m^\top \ve z_{t_m} \right) \rmd R_m(\ve z_{t_1}, \dots, \ve z_{t_m})\eqsp.
\end{eqnarray*}

We will need the following statement used in  
~\cite{gassiat:lecorff:lehericy:2020} and \cite{gassiat:lecorff:lehericy:2019}. We provide a proof at the end of the section for completeness, see also \cite{shabat1992introduction}.
\begin{Lemma}
\label{lem:analytic}
If a multivariate function is analytic on the whole multivariate complex space and is the null function on an open set of the multivariate real space or on an open set of the multivariate purely imaginary space, then it is the null function on the whole multivariate complex space. 
\end{Lemma}

Thus, equation~\eqref{eq_fn_carac_z} can be extended on $(\Cbb^M)^m$, which shows that for all $(\ve u_1, \dots, \ve u_m) \in (\Cbb^M)^m$,
\begin{equation*}
\Phi_{R_m}(\ve u_1, \dots, \ve u_m)
    = \Phi_{\tilde{R}_m}(\ve u_1, \dots, \ve u_m) \prod_{k=1}^m \exp\{Q(\ve u_k)\}\eqsp.
\end{equation*}
As $\Phi_{R_m}$ and $\Phi_{\tilde{R}_m}$ are characteristic functions, $Q$ has no constant term. The degree 1 term corresponds to a translation parameter. Without loss of generality, assume that $\ve z_{t_1}$ is centered under $R_m$ and $\tilde{R}_m$, then
\begin{equation*}
i\E_{R_m}[\ve z_{t_1}]
    = \nabla_{\ve u_1} \Phi_{R_m}(0)
    = \nabla_{\ve u_1} \Phi_{\tilde{R}_m}(0) + \nabla Q(0)
    = i\E_{\tilde{R}_m}[\ve z_{t_1}] + \nabla Q(0)\eqsp,
\end{equation*}
which entails $\nabla Q(0) = 0$. Thus, $Q$ only has terms of degree 2, which means it is a quadratic form in $\mathbb{R}^M$. Writing $Q(\ve u) = \ve u^\top(Q_+-Q_-)\ve u$ where $Q_+$ and $Q_-$ are the positive semi-definite matrices corresponding to the positive and negative eigenvalues of $Q$ respectively, yields
\begin{equation*}
\Phi_{R_m}(\ve u_1, \dots,\ve u_m) \prod_{k=1}^m \exp\left\{- \ve u_k^\top Q_+ \ve u_k\right\}
    = \Phi_{\tilde{R}_m} (\ve u_1, \dots, \ve u_m) \prod_{k=1}^m \exp\left\{-\ve u_k^\top Q_- \ve u_k\right\}\eqsp.
\end{equation*}
From this decomposition, we deduce that if $\zbf \sim R_m$, $\tilde{\zbf} \sim \tilde{R}_m$, and $(\ve v_k)_{1 \leq k \leq m}$ (resp. $(\tilde{\ve v}_k)_{1 \leq k \leq m}$) are i.i.d. multivariate Gaussian random variables with mean 0 and covariance matrices $2Q_+$ (resp. $2Q_-$) that are independent of $\zbf$ (resp. $\tilde{\zbf}$), then $(\zbf_{t_k} + \ve v_k)_{1 \leq k \leq m}$ has the same distribution as $(\tilde{\zbf}_{t_k} + \tilde{\ve v}_k)_{1 \leq k \leq m}$. In particular, the supports of the $\ve v_k$, $1 \leq k \leq m$ and of the $\tilde{\ve v}_k$, $1 \leq k \leq m$, are orthogonal.

Let $\Pi_-$ be the orthogonal projection on the support of $\tilde{\ve v}_k$, then $\Pi_- \zbf_{t_k} = \Pi_- \tilde{\zbf}_{t_k} + \tilde{\ve v}_k$, which by assumption (A3) entails $Q_- = 0$ (otherwise, take a non-zero $\eta$ in the support of $\tilde{\ve v}_k$). Since $\tilde{\zbf}$ satisfies the same assumptions as $\zbf$, $Q_+ = 0$ for the same reason. Thus, $Q = 0$, so that $\Phi_{R_m} = \Phi_{\tilde{R}_m}$, and then $R_m = \tilde{R}_m$, and likewise $P = \tilde{P}$.

\paragraph{Proof of \eqref{eq_noise}.}
Since the distribution of $(\ve x_{t_1},\ve x_{t_2})$ in the model~\eqref{eq:deconvol} is the same under $(R_2, P)$ and $(\tilde{R}_2, \tilde{P})$ (likewise for the distribution of $\ve x_{t}$ under $(R_1, P)$ and $(\tilde{R}_1, \tilde{P})$ for any $t$), we get that for all $\ve u \in \R^M$,
\begin{equation}
\label{eq:fonda0} 
\Phi_{P}(\ve u)\Phi_{R_1}(\ve u)=\Phi_{\tilde{P}}(\ve u)\Phi_{\tilde{R}_1}(\ve u)
\end{equation}
and for all $(\ve u_1, \ve u_2) \in (\R^M)^2$,
\begin{equation}
\label{eq:fonda00} 
\Phi_{P}(\ve u_1)
\Phi_{P}(\ve u_2)
\Phi_{R_2}(\ve u_1, \ve u_2)
    = \Phi_{\tilde{P}}(\ve u_1)
        \Phi_{\tilde{P}}(\ve u_2)
        \Phi_{\tilde{R}_2}(\ve u_1, \ve u_2)\eqsp.
\end{equation}
There exists a neighborhood $W$ of $0$ in $\R^M$ such that 
$\Phi_P$ and $\Phi_{\tilde{P}}$ do not vanish on $W$, so that equations~\eqref{eq:fonda0} and~\eqref{eq:fonda00} give that for all $(\ve u_1, \ve u_2) \in W^2$,
\begin{equation}
\label{eq:fonda1}    
\Phi_{R_2}(\ve u_1, \ve u_2)\Phi_{\tilde{R}_1}(\ve u_1)\Phi_{\tilde{R}_1}(\ve u_2)=\Phi_{\tilde{R}_2}(\ve u_1, \ve u_2)\Phi_{R_1}(\ve u_1)\Phi_{R_1}(\ve u_2)\eqsp.
\end{equation}
Application of Lemma~\ref{lem:analytic} yields now that \eqref{eq:fonda1} holds for all $(\ve u_1, \ve u_2) \in (\Cbb^M)^2$. Using Assumption (A2) and Lemma~\ref{lem:analytic} we easily deduce from \eqref{eq:fonda1} that the set of zeros of $\Phi_{R_1}$ and $\Phi_{\tilde{R}_1}$ are equal. Then, using Assumption (A1) and Hadamard's factorization Theorem, see \cite{Stein:complex} (Chapter 5 Theorem 5.1), and arguing variable by variable, we deduce that there exists a function $Q$ on $\Cbb^M$ such that, for all $i=1,\dots,M$, $Q$ is a polynomial function with degree at most $2$ (and coefficients depending on $(u^{(1)},\dots,u^{(i-1)},u^{(i+1)},\dots,u^{(M)})$) and for all $\ve u=(u^{(1)},\dots,u^{(M)})\in\Cbb^M$, 
\begin{equation*}
    \Phi_{R_1}(\ve u)=\Phi_{\tilde{R}_1}(\ve u)\exp(Q(\ve u)).
\end{equation*}
Using again Assumption (A1) allows to deduce that $Q$ has total degree $2$. Coming back to equation~\eqref{eq:fonda0} yields for all $\ve u \in\R^M$,
\begin{equation}
\Phi_{P}(\ve u)\Phi_{\tilde{R}_1}(\ve u)\exp(Q(\ve u))
    =\Phi_{\tilde{P}}(\ve u)\Phi_{\tilde{R}_1}(\ve u)
\end{equation}
which, on the neighborhood $V$ of $0$ in $\R^M$ where $\Phi_{\tilde{R}_1}$ does not vanish, proves \eqref{eq_noise}.

\paragraph{Proof of Lemma~\ref{lem:analytic}}

We prove the statement by induction on the number $d$ of variables. If $h$ is analytic on $\Cbb$ and is not the null function, then $h$ has isolated zeros, 
so that Lemma \ref{lem:analytic} holds for $d=1$. Assume that the lemma holds for analytic functions on $\Cbb^d$ and let $h$ be an analytic function on $\Cbb^{d+1}$ which is the null function on an open set $A$ of $\R^{d+1}$. Then, there exists open sets $B_{1},\ldots,B_{d+1}$ of $\R$ such that $B_{1}\times \cdots \times B_{d+1}\subset A$. For any $t\in B_{d+1}$, let $h_t:\Cbb^d \rightarrow \Cbb$ such that $h_{t}(\cdot)=h(\cdot,t)$, then $h_{t}$ is analytic on $\Cbb^{d}$ and is the null function on $B_{1}\times \cdots \times B_{d}$ so that by the induction hypothesis, for all $z\in\Cbb^{d}$, $h_{t}(z)=0$, that is $h(z,t)=0$  for all $z\in\Cbb$ and for all $t\in B_{d+1}$. Therefore, for any $z\in\Cbb^{d}$, the function $h(z,\cdot)$ is analytic on $\Cbb$ and is the null function on $B_{d+1}$ so that for any $z_0 \in \Cbb$, $h(z,z_0)=0$ and $h$ is the null function. The proof when $h$  is the null function on an open set of the multivariate purely imaginary space is similar.

\subsection{Proof of Theorem~\ref{theo2}}
\label{sec:proof:f:manifold}
In the following, the index $m$ may be dropped in the notations $p_m^{(i)}$ and $Q_m^{(i)}$ when there is no confusion. Let $p^{(i)}$, $\tilde{p}^{(i)}$, $\ve f$ and $\tilde{\ve f}$ be such that if $\ve s \sim p^{(i)}$ and $\tilde{\ve s} \sim \tilde{p}^{(i)}$, then $\ve f(\ve s)$ and $\tilde{\ve f}(\tilde{\ve s})$ have the same distribution.
Write $\ve g =\ve f^{-1}$ and $\tilde{\ve g} = \tilde{\ve f}^{-1}$.

Let $\ve x_1, \dots, \ve x_m \in \Mcal$. For each $k \in \{1, \dots, m\}$, let $\vartheta_k \in \Theta$ such that $\ve x_k \in U_{\vartheta_k}$ and let $\ve w_k = \varphi_{\vartheta_k}(\ve x_k)$.
Writing the density of the random vector $(\varphi_{\vartheta_1}(\ve f(\ve s_{t_1})), \ldots, \varphi_{\vartheta_m}(\ve f(\ve s_{t_m})))$ at $(\ve w_1, \dots, \ve w_m)$ with respect to the Lebesgue measure for the two parameterizations, yields
\begin{multline}
\label{eq:density:S_manifold}
\prod_{k=1}^m |J_{\ve g \circ \varphi_{\vartheta_j}^{-1}}(\ve w_k)| \prod_{i=1}^N p^{(i)}((\ve g^{(i)} \circ \varphi_{\vartheta_1}^{-1})(\ve w_1), \dots, (\ve g^{(i)} \circ \varphi_{\vartheta_m}^{-1})(\ve w_m)) \\
    = \prod_{k=1}^m |J_{\tilde{\ve g} \circ \varphi_{\vartheta_k}^{-1}}(\ve w_k)| \prod_{i=1}^N \tilde{p}^{(i)}((\tilde{\ve g}^{(i)} \circ \varphi_{\vartheta_1}^{-1})(\ve w_1), \dots, (\tilde{\ve g}^{(i)} \circ \varphi_{\vartheta_m}^{-1})(\ve w_m))\eqsp.
\end{multline}
Let $k,\ell \in \{1, \dots, m\}$ and $u,v \in \{1, \dots, N\}$ be such that $k \neq \ell$, then  by \eqref{eq:density:S_manifold},
\begin{multline*}
\sum_{i=1}^N \frac{\partial^2}{\partial w_k^{(u)} \partial w_\ell^{(v)}} \log p^{(i)}((\ve g^{(i)} \circ \varphi_{\vartheta_1}^{-1})(\ve w_1), \dots, (\ve g^{(i)} \circ \varphi_{\vartheta_m}^{-1})(\ve w_m)) \\
    = \sum_{i=1}^N \frac{\partial^2}{\partial w_k^{(u)} \partial w_\ell^{(v)}} \log \tilde{p}^{(i)}((\tilde{\ve g}^{(i)} \circ \varphi_{\vartheta_1}^{-1})(\ve w_1), \dots, (\tilde{\ve g}^{(i)} \circ \varphi_{\vartheta_m}^{-1})(\ve w_m))
    \eqsp,
\end{multline*}
that is
\begin{multline*}
\sum_{i=1}^N \frac{\partial^2 \log p^{(i)}}{\partial s_k^{(i)} \partial s_\ell^{(i)}}\left((\ve g^{(i)} \circ \varphi_{\vartheta_1}^{-1})(\ve w_1), \dots, (\ve g^{(i)} \circ \varphi_{\vartheta_m}^{-1})(\ve w_m)\right) \frac{\partial (\ve g^{(i)} \circ \varphi_{\vartheta_k}^{-1})}{\partial w^{(u)}}(\ve w_k) \frac{\partial (\ve g^{(i)} \circ \varphi_{\vartheta_\ell}^{-1})}{\partial w^{(v)}}(\ve w_\ell)
    \\
    = \sum_{i=1}^N \frac{\partial^2 \log \tilde{p}^{(i)}}{\partial s_k^{(i)} \partial s_\ell^{(i)}}\left((\tilde{\ve g}^{(i)} \circ \varphi_{\vartheta_1}^{-1})(\ve w_1), \dots, (\tilde{\ve g}^{(i)} \circ \varphi_{\vartheta_m}^{-1})(\ve w_m)\right) \frac{\partial (\tilde{\ve g}^{(i)} \circ \varphi_{\vartheta_k}^{-1})}{\partial w^{(u)}}(\ve w_k) \frac{\partial (\tilde{\ve g}^{(i)} \circ \varphi_{\vartheta_\ell}^{-1})}{\partial w^{(v)}}(\ve w_\ell)
    \eqsp.
\end{multline*}
For all $(\ve s_1, \dots, \ve s_m) \in S^m$, let
\begin{align*}
q_{i,(k,\ell)} &= \frac{\partial^2 \log p^{(i)}}{\partial s_k^{(i)} \partial s_\ell^{(i)}}\,, \quad \tilde{q}_{i,(k,\ell)} = \frac{\partial^2 \log \tilde{p}^{(i)}}{\partial s_k^{(i)} \partial s_\ell^{(i)}}\eqsp,\\
D_{k,\ell}(\ve s_1, \dots, \ve s_m) &= \text{diag}\left( q_{i,(k,\ell)}\left(s_1^{(i)}, \dots, s_m^{(i)}\right) \right)_{1 \leq i \leq N}\eqsp,\\
\tilde{D}_{k,\ell}(\ve s_1, \dots, \ve s_m) &= \text{diag}\left( \tilde{q}_{i,(k,\ell)}\left((\tilde{\ve g}^{(i)} \circ \ve g^{-1})(\ve s_1), \dots, (\tilde{\ve g}^{(i)} \circ \ve g^{-1})(\ve s_m)\right) \right)_{1 \leq i \leq N}\eqsp,
\end{align*}
so that, writing $(J_a)_{ij} = \partial a_i/\partial x_j$ the Jacobian matrix of the map $a$ and $\ve s_j = \ve g(\ve x_j)$ for each $j \in \{1, \dots, m\}$,
\begin{equation*}
J_{\ve g \circ \varphi_{\vartheta_k}^{-1}}(\ve w_k)^\top D_{k,\ell}(\ve s_1, \dots, \ve s_m) J_{\ve g \circ \varphi_{\vartheta_\ell}^{-1}}(\ve w_\ell)
    = J_{\tilde{\ve g} \circ \varphi_{\vartheta_k}^{-1}}(\ve w_k)^\top \tilde{D}_{k,\ell}(\ve s_1, \dots, \ve s_m) J_{\tilde{\ve g} \circ \varphi_{\vartheta_\ell}^{-1}}(\ve w_\ell)\eqsp.
\end{equation*}
Note that for all $\ve w \in \varphi_{\vartheta_k}(U_{\vartheta_k})$,
\begin{equation*}
J_{\tilde{\ve g} \circ \varphi_{\vartheta_k}^{-1}}(\ve w) (J_{\ve g \circ \varphi_{\vartheta_k}^{-1}}(\ve w))^{-1}
	= J_{\tilde{\ve g} \circ \ve g^{-1}}( (\ve g \circ \varphi_{\vartheta_k}^{-1})(\ve w) )\eqsp,
\end{equation*}
so that for all $(\ve s_1, \dots, \ve s_m) \in S^m$,
\begin{equation}
\label{eq:diagonalisation:D}
D_{k,\ell}(\ve s_1, \dots, \ve s_m)
    = J_{\tilde{\ve g} \circ \ve g^{-1}}(\ve s_k)^\top \tilde{D}_{k,\ell}(\ve s_1, \dots, \ve s_m) J_{\tilde{\ve g} \circ \ve g^{-1}}(\ve s_\ell)\eqsp.
\end{equation}

Consider the following assertion. 
\begin{itemize}
\item (P) For all $\ve s$ in a dense subset of $S$, there exist integers $k,\ell \in \{1, \dots, m\}$ with $k \neq \ell$ and $\ve s_1, \dots, \ve s_{k-1},$ $\ve s_{k+1}, \dots, \ve s_{m} \in S$ such that all entries of the vector
\begin{equation*}
\left(
\frac{
        q_{i,(k,\ell)}(\dots, s^{(i)}, \dots, s^{(i)}, \dots)
        q_{i,(k,\ell)}(\dots, s_\ell^{(i)}, \dots, s_\ell^{(i)}, \dots)
    }{
        q_{i,(k,\ell)}(\dots, s^{(i)}, \dots, s_\ell^{(i)}, \dots)^2
    }
\right)_{1 \leq i \leq N}
\end{equation*}
are distinct ($s^{(i)}$ and $s_\ell^{(i)}$ are in the positions $k$ and $\ell$ in the equation above).
\end{itemize}
Assume that (P) holds. [We shall prove below that (P) holds under the assumptions of Theorem~\ref{theo2}]. Let $\ve s = (\ve s_1, \dots, \ve s_m) \in S$ such that $D_{k,\ell}(\ve s_1, \dots, \ve s_m)$ is invertible (any $\ve s$ in a dense subset of $S$ works thanks to assumption B1).
For ease of notation in the following sequence of equations, we drop all unused subscripts and parameters, thus writing $J(\ve s_k)$ instead of $J_{\tilde{\ve g} \circ \ve g^{-1}}(\ve s_k)$ and $D(\ve s_k, \ve s_\ell)$ instead of $D_{k,\ell}(\ve s_1, \dots, \ve s_k, \dots, \ve s_\ell, \dots, \ve s_m)$ (and likewise for $\tilde J$ and $\tilde D$). We follow the arguments of the proof of Lemma 2 in \cite{Hyva17AISTATS} to deduce from~\eqref{eq:diagonalisation:D} an eigenvalue decomposition. Write~\eqref{eq:diagonalisation:D} for several parameters:
\begin{align*}
D(\ve s_k, \ve s_k) &= J(\ve s_k)^\top \tilde{D}(\ve s_k, \ve s_k) J(\ve s_k), \\
D(\ve s_k, \ve s_\ell) &= J(\ve s_k)^\top \tilde{D}(\ve s_k, \ve s_\ell) J(\ve s_\ell) \\
    &= J(\ve s_\ell)^\top \tilde{D}(\ve s_k, \ve s_\ell) J(\ve s_k) \quad \text{by symmetry}, \\
D(\ve s_\ell, \ve s_\ell) &= J(\ve s_\ell)^\top \tilde{D}(\ve s_\ell, \ve s_\ell) J(\ve s_\ell),
\end{align*}
which altogether entails
\begin{multline*}
D(\ve s_k, \ve s_\ell)^{-1}
    D(\ve s_\ell, \ve s_\ell)
    D(\ve s_k, \ve s_\ell)^{-1}
    D(\ve s_k, \ve s_k) \\
= J(\ve s_k)^{-1} \left[ 
    \tilde{D}(\ve s_k, \ve s_\ell)^{-1}
    \tilde{D}(\ve s_\ell, \ve s_\ell)
    \tilde{D}(\ve s_k, \ve s_\ell)^{-1}
    \tilde{D}(\ve s_k, \ve s_k)
\right] J(\ve s_k).
\end{multline*}
The vector in assertion (P) contains the diagonal entries of this diagonal matrix. If they are all distinct, the eigenvalue decomposition is unique, which means that $J(\ve s_k)$ is the product of a permutation matrix and a diagonal matrix.


Thus, $J_{\tilde{\ve g} \circ \ve g^{-1}}$ is the product of a permutation matrix with a diagonal matrix on a dense subset of $S$, and hence on $S$ by regularity of $\ve g$ and $\tilde{\ve g}$.

For any permutation matrix $P$, the set of all $\ve s \in S$ where $J_{\tilde{\ve g} \circ \ve g^{-1}}(\ve s)$ is the product of $P$ with an invertible diagonal matrix $D(\ve s)$ is both open (by continuity of $J_{\tilde{\ve g} \circ \ve g^{-1}}$) and closed (if $\ve s_n \rightarrow \ve s$ are such that $J_{\tilde{\ve g} \circ \ve g^{-1}}(\ve s_n) = P D_n$ for all $n$, then by continuity the permutation matrix at $\ve s$ is also $P$ and since the jacobian is always invertible by the diffeomorphism assumption, $\lim_n D_n$ exists and is invertible). Thus, by connexity of $S$, the permutation is the same for all $\ve s \in S$. For the next paragraph, we assume without loss of generality that it is the identity permutation.

Therefore, since for all $j$ and $s^{(j)} \in S^{(j)}$, the set $S^{(1)} \times \dots \times S^{(j-1)} \times \{s_j\} \times S^{(j+1)} \times \dots \times S^{(N)}$ is connected, $(\tilde{\ve g} \circ \ve g^{-1})^{(j)}$ is constant on this set, and thus it depends on $s^{(j)}$ only. It is bijective on $S^{(j)}$ because both $\ve g$ and $\tilde{\ve g}$ are. Thus, $\ve g = \tilde{\ve g}$ up to a permutation of the coordinates and a bijective transformation of each coordinate.

Let us now prove that assertion (P) is true. 
The negation of (P) is that 
there exists an open set $A \subset S$ such that for all $\ve s \in A$, for all $k,\ell \in \{1, \dots, m\}$ with $k \neq \ell$ and for all $(\ve s_1, \dots, \ve s_{k-1}, \ve s_{k+1}, \dots, \ve s_{m}) \in S^{m-1}$, there exists $i, j \in \{1, \dots, N\}$ with $i \neq j$ such that
\begin{multline}
\label{eq_egalite_ratio_q}
\frac{
        q_{i,(k,\ell)}(\dots, s^{(i)}, \dots, s^{(i)}, \dots)
        q_{i,(k,\ell)}(\dots, s^{(i)}_\ell, \dots, s^{(i)}_\ell, \dots)
    }{
        q_{i,(k,\ell)}(\dots, s^{(i)}, \dots, s^{(i)}_\ell, \dots)^2
    }
    \\
    = \frac{
        q_{j,(k,\ell)}(\dots, s^{(j)}, \dots, s^{(j)}, \dots)
        q_{j,(k,\ell)}(\dots, s^{(j)}_\ell, \dots, s^{(j)}_\ell, \dots)
    }{
        q_{j,(k,\ell)}(\dots, s^{(j)}, \dots, s^{(j)}_\ell, \dots)^2
    }
    \eqsp.
\end{multline}
Let $\ve s \in A$, $k,\ell \in \{1, \dots, m\}$ with $k \neq \ell$. For all $(i,j)\in\{1,\ldots,N\}^2$ with $i\neq j$, define $\tilde{S}_{i,j}$ the subset of $S^{m-1}$ such that for all $(\ve s_1, \dots, \ve s_{k-1}, \ve s_{k+1}, \dots, \ve s_m) \in \tilde{S}_{i,j}$, equation~\eqref{eq_egalite_ratio_q} holds. Since the sets $\tilde{S}_{i,j}$, $(i,j)\in\{1,\ldots,N\}^2$, $i\neq j$, form a partition of $S^{m-1}$, which has non-empty interior, there exists at least one pair $(i,j)$ such that the closure of $\tilde{S}_{i,j}$ contains a non-empty open subset $O_{i,j}$.
Since $q_{i,(k,\ell)}$ and $q_{j,(k,\ell)}$ are non zero almost everywhere by the uniform $(k,\ell)$-dependency assumption, we may assume without loss of generality that the denominators of equation~\eqref{eq_egalite_ratio_q} are non zero for all $(\ve s_1, \dots, \ve s_{k-1}, \ve s_{k+1}, \dots, \ve s_m) \in O_{i,j}$.
Thus, by continuity of $q_{i,(k,\ell)}$ and $q_{j,(k,\ell)}$, the terms of equation~\eqref{eq_egalite_ratio_q} do not depend on the choice of element in $O_{i,j}$: write $f_{i,(k,\ell)}(s^{(i)}, O_{i,j})$ the left hand term and $f_{j,(k,\ell)}(s^{(j)}, O_{i,j})$ the right hand term.

Let $k,\ell \in \{1, \dots, m\}$ with $k \neq \ell$. Let $(V_n)_{n \geq 1}$ be a basis of open sets of $(\R^N)^{m-1}$. For all $(i,j) \in \{1, \dots, N\}$ with $i \neq j$ and $n \in \Nbb^*$, let $A_{(i,j),n}$ be the subset of $A$ such that for all $\ve s \in A_{(i,j),n}$ and all $(\ve s_1, \dots, \ve s_{k-1}, \ve s_{k+1}, \dots, \ve s_m) \in V_n$, equation~\eqref{eq_egalite_ratio_q} holds. Then, $A = \bigcup_{n \geq 1} \bigcup_{i \neq j} A_{(i,j),n}$ (since $O_{i,j}$ contains at least one of the sets of the basis $(V_n)_{n \geq 1}$) and thus there exists $i \neq j$ and $n$ such that the interior of the closure of $A_{(i,j),n}$ is non-empty (otherwise $A$ would be a meagre set and thus have empty interior by Baire's category theorem, which is absurd since $A$ is a non-empty open set). Let $i,j,n$ be such that the closure of $A_{(i,j),n}$ has non-empty interior, and $B$ be a non-empty subset of the closure of $A_{(i,j),n}$.
Since $q_{i,(k,\ell)}$ and $q_{j,(k,\ell)}$ are non zero almost everywhere by the uniform $(k,\ell)$-dependency assumption, we may take an open set $V \subset V_n$ and assume without loss of generality that the denominators of equation~\eqref{eq_egalite_ratio_q} are non zero for all $(\ve s_1, \dots, \ve s_{k-1}, \ve s_{k+1}, \dots, \ve s_m) \in V$ and all $\ve s \in B$.
Thus, by continuity of $q_{i,(k,\ell)}$ and $q_{j,(k,\ell)}$, the terms of equation~\eqref{eq_egalite_ratio_q} do not depend on the choice of element in $B$ or $V$.

To summarize, this means that for all $k,\ell \in \{1, \dots, m\}$ with $k \neq \ell$, there exists $(i,j) \in \{1, \dots, N\}$ with $i \neq j$, a constant $c$ and an open set $A' \subset S^{m}$ such that for all $\ve s = (\ve s_1, \dots, \ve s_m) \in A'$,
\begin{align*}
\label{eq_egalite_ratio_q}
q_{i,(k,\ell)}(\dots, s^{(i)}_k, \dots, s^{(i)}_\ell, \dots)^2
    = c q_{i,(k,\ell)}(\dots, s^{(i)}_k, \dots, s^{(i)}_k, \dots)
        q_{i,(k,\ell)}(\dots, s^{(i)}_\ell, \dots, s^{(i)}_\ell, \dots)\eqsp,
\\
q_{j,(k,\ell)}(\dots, s^{(j)}_k, \dots, s^{(j)}_\ell, \dots)^2
    = c q_{j,(k,\ell)}(\dots, s^{(j)}_k, \dots, s^{(j)}_k, \dots)
        q_{j,(k,\ell)}(\dots, s^{(j)}_\ell, \dots, s^{(j)}_\ell, \dots)
    \eqsp.
\end{align*}
This situation is excluded by the local $(k,\ell)$-non quasi Gaussianity assumption, therefore the negation of (P) is false, therefore $\ve g = \tilde{\ve g}$ up to permutation and bijective transformation of each coordinate.

\subsection{Proof of Theorem~\ref{th:A2}}

For all $\eta \in \Cbb^m$,
\begin{multline*}
\mathbb{E}\left[\exp \left\{\langle \eta, \ve z_{t_2}\rangle \right\} \middle| \; \ve z_{t_1}\right]
	= \frac{\sum_{u, v} \pi(u) Q(u, v) \gamma_{u}(\ve z_{t_1}) \int \exp(\langle \eta, \ve z \rangle) \gamma_{v}(\ve z) d\ve z}
		{\sum_{u} \pi(u) \gamma_{u}(\ve z_{t_1})} \\
	=\frac{\sum_{u}\alpha_{u}(\eta) \pi(u) \gamma_{u}(\ve z_{t_1})}{\sum_{u} \pi(u) \gamma_{u}(\ve z_{t_1})}\,,
\end{multline*}
with $\alpha_{u}(\eta) = \sum_{v} Q(u, v) \int \exp(\langle \eta, \ve z \rangle) \gamma_{v}(\ve z)d\ve z$.

Assume that the emission densities $(\gamma_u)_{1 \leq u \leq K}$ are linearly independent and $\pi(u) > 0$ for all $u \in \{1, \dots, K\}$, then the only situation where $\E[\exp\{\langle \eta, \ve z_{t_2} \rangle\} | \ve z_{t_1}]$ is the null random variable is when $\alpha_u(\eta) = 0$ for all $u \in \{1, \dots, K\}$.
If the functions $(\eta \mapsto \int \exp(\langle \eta, \ve z\rangle)\gamma_{v}(\ve z)d\ve z)_{1 \leq v \leq K}$ do not have simultaneous zeros and $Q$ has full rank, this is not possible.

\subsection{Proof of Theorem~\ref{th:simplehmm}}

We prove that the result holds for all $i=1,\ldots,N$ and drop the index $i$ in this proof for ease of notation. Denote by \begin{equation*}
	\label{eq:1}
	\Lambda \coloneqq
	\begin{pmatrix}
		1 - p & p\\
		q & 1 - q
	\end{pmatrix}%
\end{equation*}
the transition matrix of the hidden chain. Then, the stationary distribution is given by $\pi (0)=q/(p+q)$, $\pi(1)= p/(p+q)$, and the distribution of $2$ consecutive observations is given by, for all $(a,b)$ in the support:
$$
p_{2}(a,b)=\frac{q(1-p)}{p+q}\gamma_{0}(a)\gamma_{0}(b)+
\frac{qp}{p+q}\gamma_{0}(a)\gamma_{1}(b)+
\frac{pq}{p+q}\gamma_{1}(a)\gamma_{0}(b)+
\frac{p(1-q)}{p+q}\gamma_{1}(a)\gamma_{1}(b)\eqsp.
$$
If $Q_{2}=\log p_{2}$ then simple computations lead to
$$
(p+q)^2 p_{2}(a,b)^{2}\frac{\partial^2 Q_{2}}{\partial a \partial b}=pq(1-p-q)(\gamma_{0}(a)\gamma'_{1}(a)-\gamma'_{0}(a)\gamma_{1}(a))(\gamma_{0}(b)\gamma'_{1}(b)-\gamma'_{0}(b)\gamma_{1}(b))\eqsp.
$$
Since $\gamma_{0}(a)\gamma'_{1}(a)-\gamma'_{0}(a)\gamma_{1}(a)=0$ for $a$ in an open subset of the support if and only if on this interval $\gamma_0$ and $\gamma_1$ are proportional, assumption (B1) is satisfied if and only if on any open interval $\gamma^{(i)}_0$ and $\gamma^{(i)}_1$ are not proportional. Moreover, on the set of couples $(a,b)$ such that $\frac{\partial^2 Q_{2}}{\partial a \partial b}\neq 0$, 
$$
\log \left(\frac{\partial^2 Q_{2}}{\partial a \partial b}\right)=\log[|pq(1-p-q)|]-2\log(p+q)-2\log p_{2}(a,b)+h(a)+h(b)\eqsp,
$$
where $h(a)=|\gamma_{0}(a)\gamma'_{1}(a)-\gamma'_{0}(a)\gamma_{1}(a)|$. We deduce easily that (B2) is satisfied if and only if on any open interval $\gamma^{(i)}_0$ and $\gamma^{(i)}_1$ are not proportional.
\section{Identifiability in Gaussian case} \label{gaussiandiscussion}

Theorem~\ref{theo2} has a condition on "non-quasi-Gaussianity" which is a generalization of the property of non-Gaussianity typical in ICA. Here, we consider the case of Gaussian noise-free data. Separation is actually possible by the temporal dependencies, but under a stricter condition. We put together results by \citet{Hyva17AISTATS} and \citet{Belouchrani97}, and arrive at the following result:
\begin{Theorem} \label{th:gaussian}
Assume the data follows the noise-free mixing model $\ve x_t=\ve f(\ve s_t)$ where $\ve s_t$ is a Gaussian process with independent components, and $\ve f$ is a ${\cal C}^2$ diffeomorphism with $M=N$. Assume further that
\begin{itemize} \item The autocovariance functions $c_i(\tau)=\text{cov}(s_t^{(i)},s_{t-\tau}^{(i)})$ are all distinct (i.e.\ any two of them for $i,i'$ are not equal). (Here, $\tau$ takes values in the set allowed by the definition of the index set.)
\end{itemize}
Then, $\ve f^{-1}$ and $\ve f$ can be recovered up to permutation and coordinate-wise linear transformations (applied on the components $\ve s_t^{(i)}$) from the distribution of $\ve x_t$.
\end{Theorem}
The proof is a straightforward implication of two theorems proven earlier: The nonlinear part is identifiable according to Theorem~2 by \citet{Hyva17AISTATS} but a linear indeterminacy remains; here we need to note that $\bar{\alpha}$ in \citep{Hyva17AISTATS} is a linear function for a Gaussian process. Subsequently the linear part can be identified, thanks to the autocovariance assumption above, as in  Theorem~2 of \citet{Belouchrani97}. 

Note that in the Gaussian case, it is not possible to apply Theorem~\ref{theo1} since (A3) cannot hold. Thus, Theorem~\ref{th:gaussian} only applies for noise-free data.

\section{Learning and inference for \modelname} \label{sec:apdx:estimation}
The \modelname generative model, as introduced in Section~\ref{sec:sld-snica} can be written as:
\begin{align}
    p(u_1^{(i)}) &= \prod_{k=1}^K (\pi_k^{(i)})^{\delta(u_1^{(i)} = k)} \\
    p(u_t^{(i)}\mid u_{t-1}^{(i)}) &= \prod_{k=1}^K\prod_{\ell=1}^K (A_{k\ell}^{(i)})^{ \delta(u_t^{(i)}=k)\delta(u_{t-1}^{(i)}=\ell)} \\
    p(\ve y_1^{(i)}\mid u_1^{(i)}) &= \prod_{k=1}^K\mathcal{N}(\ve y_1^{(i)} ; \bar{\ve b}_k^{(i)}, \bar{\m Q}_k^{(i)})^{\delta(u_1^{(i)}=k)} \\
    p(\ve y_t^{(i)}\mid \ve y_{t-1}^{(i)}, u_t^{(i)}) &= \prod_{k=1}^K \mathcal{N}(\ve y_t^{(i)} ; \m B_k^{(i)} \ve y_{t-1}^{(i)} + \m b_k^{(i)}, \m Q_k^{(i)})^{\delta(u_t^{(i)}=k)} \\
    p(\ve x_t\mid \ve s_t) &= \mathcal{N}(\ve x_t ; \ve f( \ve s_t), \m R)
\end{align}
where the superscript $(i)$ again denotes that each independent component $i \in \{1,\dots, N\}$ follows its own switching linear dynamical system. Also, as explained in Section \ref{sec:sld-snica}, each independent component is part of a higher dimensional latent component $\ve y_t^{(i)} = (s_t^{(i)}, y_{t, 2}^{(i)},\dots, y_{t, d}^{(i)})$. The mixing function $\ve f$ and other variables are defined as in the main text. The log-joint $\log\mathcal{L}=\log p(\ve x_{1:T}^{(1:N)}, \ve y_{1:T}^{(1:N)}, u_{1:T}^{(1:N)})$ can be written as:
\begin{align}
	\log \mathcal{L} &= \sum_{t=1}^T \log p(\ve x_t \mid \ve s_t) + \sum_{i=1}^N\left(\log p(u_1^{(i)}) +\log p(\ve y_1^{(i)} \mid u_1^{(i)}) \right. \nonumber \\
	&\left. \sum_{t=2}^T \log p(u_t^{(i)}\mid u_{t-1}^{(i)}) + \log p(\ve y_t^{(i)} \mid \ve y_{t-1}^{(i)}, u_t^{(i)}) \right) \eqsp.
\end{align}
The marginal likelihood is intractable and hence we instead optimize the variational evidence lower bound (ELBO), denoted here $\log \widehat{\mathcal{L}}$, under the assumption that the posterior factorizes as per \begin{align}
	q(\ve y_{1:T}^{(1:N)}, u_{1:T}^{(1:N)}) = \prod_{i=1}^N q(\ve y_{1:T}^{(i)}) q(u_{1:T}^{(i)}). \label{eq:factorization}
\end{align}
The ELBO can thus be written as:
\begin{align}
    \log \widehat{\mathcal{L}} &= \E_{q}\bigg[\log \frac{p(\ve x_{1:T}, \ve y_{1:T}^{(1:N)}, u_{1:T}^{(1:N)})}{q(\ve y_{1:T}^{(1:N)}, u_{1:T}^{(1:N)})} \bigg] \nonumber\\
    &= \E_{q}\bigg[\sum_{t=1}^T\log p(\ve x_t\mid\ve s_t^{(1)}, ..., \ve s_t^{(N)}) + \sum_{i=1}^N\log\frac{p(\ve y_{1:T}^{(i)}\mid u_{1:T}^{(i)}) p(u_{1:T}^{(i)})}{q(\ve y_{1:T}^{(i)})q(u_{1:T}^{(i)})} \bigg] \nonumber\\
    &= \E_{q}\bigg[\sum_{t=1}^T\log p(\ve x_t\mid\ve s_t^{(1)}, ..., \ve s_t^{(N)})\bigg] + \sum_{i=1}^N\Bigg(-\mathrm{KL}\bigg[q(u_{1:T}^{(i)})\bigg| p(u_{1:T}^{(i)})\bigg] + \mathrm{H}\bigg[q(\ve y_{1:T}^{(i)})\bigg] \nonumber\\    
    &\qquad+ \E_{q}\bigg[\log p(\ve y_{1:T}^{(i)}\mid u_{1:T}^{(i)})\bigg]\Bigg) \nonumber\\
    &= \E_{q}\bigg[\sum_{t=1}^T\log p(\ve x_t\mid\ve s_t^{(1)}, ..., \ve s_t^{(N)})\bigg] + \sum_{i=1}^N\Bigg(- \mathrm{KL}\bigg[q(u_{1:T}^{(i)})\bigg| p(u_{1:T}^{(i)})\bigg] + \mathrm{H}\bigg[q(\ve s_{1:T}^{(i)})\bigg] \nonumber\\
    &\qquad+ \E_{q}\bigg[\log p(\ve s_1^{(i)}\mid u_1^{(i)}) \bigg] + \sum_{t=2}^T \E_{q}\bigg[\log p(\ve s_t^{(i)}\mid\ve s_{t-1}^{(i)}, u_t^{(i)}) \bigg]\Bigg) \label{eq:ELBO}
\end{align}
where $\mathrm{H}$ denotes Gaussian differential entropy, and $q$ is always with respect to the relevant variables. As long as all the distributions are conjugate-exponential families, we can use the Structured VAE \citet{johnson2017composing} framework for inference and learning. We provide further detail on these two steps below.

\paragraph{Inference}
Notice that we can write the latent variable part of our generative model in the following useful exponential family forms:  
\begin{align}
p(u_1^{(i)}) &= \prod_{k=1}^K \pi_k^{(i)^{\delta( u_1^{(i)} = k)}} = \exp \left\{{\sum_{i=1}^K \delta( u_1^{(i)} = k)\log \pi_k^{(i)}}\right\} \nonumber = \exp \left\{\langle \ve \eta_{\ve \pi}^{(i)},  \ve \delta_{u_1}^{(i)}  \rangle\right\} \nonumber \\
p(u_t^{(i)}\mid u_{t-1}^{(i)}) &= \prod_{k=1}^K\prod_{\ell=1}^K A_{k\ell}^{(i)^{\delta(u_{t-1}^{(i)}=k)\delta(u_t^{(i)}=\ell)}} = \exp \left\{\langle \ve \eta_{\m A}^{(i)}, \ve \delta_{u_{t-1}, u_{t}}^{(i)} \rangle\right\} \\ \nonumber
p(\ve y_1^{(i)}\mid u_1^{(i)}) &= \prod_{k=1}^K\mathcal{N}(\ve y_1^{(i)} ; \bar{\ve b}_k^{(i)}, \bar{\m Q}_k^{-1^{(i)}})^{\delta(u_1^{(i)}=k)} \nonumber \\
				       &= \exp\left\{\sum_{k=1}^K \delta(u_1^{(i)}=k) \left(\langle \ve h_{1,k}^{(i)}, \ve y_1^{(i)}\rangle + \ve y_1^{(i)^T} \m J_{1, k}^{(i)} \ve y_1^{(i)} -\log Z_{1, k}^{(i)}\right) \right\} \nonumber \\
	\ve h_{1,k}^{(i)} &=\bar{\m Q}_k^{(i)}\bar{\ve b}_k^{(i)} \nonumber\\ 
	\m  J_{1, k}^{(i)} &= -\frac{1}{2} \bar{\m Q}_k^{(i)} \eqsp, \nonumber
\end{align}
where $\log Z_{1, k}^{(i)}$ is the log-normalizer, and similarly 
\begin{align}
	p(\ve y_t^{(i)}\mid \ve y_{t-1}^{(i)}, u_t^{(i)}) &= \prod_{k=1}^K \mathcal{N}(\ve y_t^{(i)} ; \m B_k^{(i)} \ve y_{t-1}^{(i)} + \m b_k^{(i)}, \m Q_k^{-1^{(i)}})^{\delta(u_t^{(i)}=k)} \nonumber \\
							  &= \exp\left\{\sum_{k=1}^K  \delta(u_t^{(i)}=k) \left(\left\langle \ve h_{k}^{(i)},  \ve y_{t-1,t}^{(i)} \right\rangle +  \ve y_{t-1,t}^{(i)^T}  \m J_{k}^{(i)} \ve y_{t-1,t}^{(i)}-\log Z_{k}^{(i)}\right) \right\} \nonumber \\
	\ve y_{t-1,t}^{(i)} &= (\ve y_{t-1}^{(i)}, \ve y_t^{(i)})^T \nonumber \\
	\ve h_{k}^{(i)} &= \begin{pmatrix}  \m B_k^{(i)^T} \m Q_k^{(i)} \m B_k^{(i)} & -\m B_k^{(i)^T} \m Q_k^{(i)}  \\ -\m Q_k^{(i)} \m B_k^{(i)} & \m Q_k^{(i)}  \end{pmatrix} \begin{pmatrix} \ve 0 \\ \ve b_k^{(i)} \end{pmatrix} \nonumber \\
	\m J_{k}^{(i)} &=-\frac{1}{2}\begin{pmatrix}  \m B_k^{(i)^T} \m Q_k^{(i)} \m B_k^{(i)} & -\m B_k^{(i)^T} \m Q_k^{(i)}  \\ -\m Q_k^{(i)} \m B_k^{(i)} & \m Q_k^{(i)}  \end{pmatrix}  \eqsp. \nonumber
\end{align}
Applying standard results from structured mean-field inference, the updates for the approximate posterior of the HMM latent variables is as follows:
\begin{align}
	q(u_{1:T}^{(i)})  &\propto \exp\left\{\log p(u_1^{(i)})+ \sum_{t=2}^T \log p(u_t^{(i)}\mid u_{t-1}^{(i)}) \right. \nonumber \\
	&+ \left. \E_{q(\ve y_{1}^{(i)})}\left[\log p(\ve y_1^{(i)}\mid u_1^{(i)})\right] + \E_{q(\ve y_{t-1,t}^{(i)})}\left[\log p(\ve y_t^{(i)}\mid \ve y_{t-1}^{(i)}, u_t^{(i)}) \right]  \right\} \nonumber \eqsp. 
\end{align}
And by plugging in the distributions explicitly gives
\begin{align}
	q(u_{1:T}^{(i)})  &\propto \exp\left\{ \langle \ve \eta_{\ve \pi^{(i)}},   \ve \delta_{u_1}^{(i)}  \rangle +  \langle  \ve \delta_{u_1}^{(i)}, \ve \rho_1^{(i)} \rangle   + \sum_{t=2}^T  \langle \ve \eta_{\m A^{(i)}}, \vect
	\left(\ve \delta_{u_{t-1}}^{(i)} \ve \delta_{u_t}^{(i)^T}\right)  \rangle +   \langle  \ve \delta_{u_t}^{(i)}, \ve \rho_t^{(i)} \rangle  \right\} \eqsp,  \label{eq:u_update}
\end{align}
where we have defined
\begin{align}
	\E_{q(\ve y_{t-1,t}^{(i)})}\left[\log p(\ve y_t^{(i)}\mid \ve y_{t-1}^{(i)}, u_t^{(i)}) \right] &= \sum_{k=1}^K  \delta(u_t^{(i)}=k) \E_{q(\ve y_{t-1,t}^{(i)})}\left[\left\langle \ve h_{t, k}^{(i)},  \ve y_{t-1,t}^{(i)} \right\rangle + \right. \nonumber \\
	& \left. \ve y_{t-1,t}^{(i)^T}  \m J_{t, k}^{(i)} \ve y_{t-1,t}^{(i)}-\log Z_{t, k}^{(i)}\right] \nonumber \\
	&=\langle \ve \delta_{u_t}^{(i)},  \ve \rho_t^{(i)} \rangle \nonumber \eqsp.
\end{align}
Equation \eqref{eq:u_update} can be viewed as a factor graph of unnormalized potentials -- we can therefore use standard message passing algorithms for efficient inference. For instance, the forward-pass is:
\begin{align}
	\alpha(u_t^{(i)}) = \sum_{u_{t-1}} \exp\left\{  \sum_{t=2}^T  \langle \ve \eta_{\m A^{(i)}}, \vect\left(\ve \delta_{u_{t-1}}^{(i)} \ve \delta_{u_t}^{(i)^T}\right)  \rangle +   \langle  \ve \delta_{u_t}^{(i)}, \ve \rho_t^{(i)} \rangle   \right\}\alpha(u_{t-1}^{(i)}) \eqsp.
\end{align}
Similarly, the standard mean-field updates for the dynamical system latent variables gives:
\begin{align}
	q(\ve y_{1:T}^{(i)}) &\propto \exp\left\{\sum_{t=1}^T \E_{\prod_{j=1}^{N\setminus i} q(\ve y_{t}^{(j)})}\left[\log p(\ve x_t\mid \ve s_t)\right]+ \E_{q(u_1^{(i)})}\left[\log p(\ve y_1^{(i)}\mid u_1^{(i)})\right] \right. \nonumber \\
			     &\left. + \sum_{t=2}^T \E_{q(u_t^{(i)})}\left[ \log p(\ve y_t^{(i)}\mid \ve y_{t-1}^{(i)}, u_t^{(i)})\right]  \right\} \eqsp. \label{eq:lds_post}
\end{align}
The problem here is that we would like to write all the factors in terms of $\ve s_t$ and $\ve y_t$ conditional on $\ve x_t$. However, due to the nonlinear mixing function, we can't write this directly in conjugate exponential family form. To resolve this, we follow \citet{johnson2017composing} and use a decoder neural network to predict approximate natural parameters such that they are in conjugate form, namely:
\begin{align}
	\E_{\prod_{N \setminus i} q(\ve y_{t}^{(j)})}\left[\log p(\ve x_t\mid \ve s_t)\right] &\propto \langle \ve v_t(\ve x_t; \ve \phi) , \ve s_t \rangle + \ve s_t^T \m W_t(\ve x_t; \ve \phi) \ve s_t \eqsp, \nonumber 
\end{align}
where $\ve v_t, \ve W_t$ are thus the outputs of the decoder network, with the latter term assumed to have diagonal structure with negative entries to ensure it's an appropriate Gaussian natural parameter. Further, due to the factored approximation assumption over $\ve y_t^{(1)},\dots, \ve y_t^{(N)}$ and thus $\ve s_t^{(1)}, \dots, \ve s_t^{(N)}$, above can be written as:

\begin{align}
 \E_{\prod_{N \setminus i} q(\ve y_{t}^{(j)})}\left[\log p(\ve x_t\mid \ve s_t)\right] &\propto  \nonumber \left(v_{t,i}+ 2\sum_{j\setminus i}^N w_{t,j,i} \E_{q(\ve y_{t}^{(j)})}\left[y_{t,1}^{(j)}\right]\right)y_{t,1}^{(i)} + w_{t,i,i}y_{t, 1}^{(i)^2} \nonumber \\
											     &= \langle \tilde{\ve v_t}^{(i)}, \ve y_t^{(i)} \rangle + \ve y_t^{(i)^T} \widetilde{\m W}^{(i)} \ve y_t^{(i)} 
\end{align}
where $\tilde{\ve v_t}^{(i)}, \widetilde{\m W}^{(i)}$ are zero everywhere except in their first indices. The other expectations in Equation \eqref{eq:lds_post} are just responsibility weighted natural parameters. For instance:
\begin{align}
	\E_{q(u_t^{(i)})}\left[ \log p(\ve y_t^{(i)}\mid \ve y_{t-1}^{(i)}, u_t^{(i)})\right] &\propto \sum_{k=1}^K  \E_{q(u_t^{(i)})}\left[ \delta(u_t^{(i)}=k)\right] \left(\left\langle \ve h_{t, k}^{(i)},  \ve y_{t-1,t}^{(i)} \right\rangle +  \ve y_{t-1,t}^{(i)^T}  \m J_{t, k}^{(i)} \ve y_{t-1,t}^{(i)}\right) \nonumber \\
											      &\propto \left\langle \tilde{\ve h}_{t}^{(i)},  \ve y_{t-1,t}^{(i)} \right\rangle +  \ve y_{t-1,t}^{(i)^T}  \tilde{\m J}_{t}^{(i)} \ve y_{t-1,t}^{(i)} \nonumber \\
	\tilde{\ve h}_{t}^{(i)} &= \sum_{k=1}^K \E_{q(u_t^{(i)})}\left[ \delta(u_t^{(i)}=k)\right] \ve h_{t,k}^{(i)} \nonumber \\
	\tilde{\m J}_{t}^{(i)} &= \sum_{k=1}^K \E_{q(u_t^{(i)})}\left[ \delta(u_t^{(i)}=k)\right] \m J_{t, k}^{(i)} \nonumber
\end{align}
The approximate posterior in \eqref{eq:lds_post} can therefore be written as:
\begin{align}
	q(\ve y_{1:T}^{(i)}) &\propto \exp\left\{\langle \tilde{\ve v}_1^{(i)}, \ve y_1^{(i)} \rangle + \ve y_1^{(i)^T} \widetilde{\m W}^{(i)} \ve y_1^{(i)} + \langle \tilde{\ve h}_{1}^{(i)}, \ve y_1^{(i)}\rangle + \ve y_1^{(i)^T} \tilde{\m J}_{1}^{(i)} \ve y_1^{(i)}\right. \nonumber \\
			     &\left. + \sum_{t=2}^T  \langle \tilde{\ve v}_t^{(i)}, \ve y_t^{(i)} \rangle + \ve y_t^{(i)^T} \widetilde{\m W}^{(i)} \ve y_t^{(i)} + \left\langle \tilde{\ve h}_{t}^{(i)},  \ve y_{t-1,t}^{(i)} \right\rangle +  \ve y_{t-1,t}^{(i)^T}  \tilde{\m J}_{t}^{(i)} \ve y_{t-1,t}^{(i)}  \right\}.
\end{align}
This can again be viewed as a factor graph on which to perform message passing. The initial forward message is
\begin{align}
	\alpha(\ve y_1) &=\exp\left\{\langle \tilde{\ve v}_1+ \tilde{\ve h}_{1}, \ve y_1 \rangle + \ve y_1^{T} \left(\widetilde{\m W}^{}+ \tilde{\m J}_{1}\right) \ve y_1  \right\}, \nonumber \\
			&=\exp\left\{\langle \ve \eta_{1}, \ve y_1 \rangle + \ve y_1^{T} \m P_1 \ve y_1  \right\}, \nonumber
\end{align}
which is an unnormalized Gaussian distribution, and we have dropped superscripts for convenience. The forward equations can be derived as follows, shown here for $t-1=1,t=2$:
\begin{align}
	\alpha(\ve y_2) &= \exp\{ \langle \tilde{\ve v}_2^{}, \ve y_2^{} \rangle + \ve y_2^{T} \widetilde{\m W} \ve y_2 \} \int_{\ve y_1}\exp\left\{ \left\langle \tilde{\ve h}_{2},  \ve y_{1,2} \right\rangle +  \ve y_{1,2}^{T}  \tilde{\m J}_{2} \ve y_{1,2} + \langle \ve \eta_{1}, \ve y_1 \rangle + \ve y_1^{T} \m P_1 \ve y_1  \right\} \nonumber.
\end{align}
Define $\ve \eta_2^{*} = (\tilde{\ve h}_{2}^1 + \ve \eta_1, \tilde{\ve h}_{2}^2)^T$ and $\m P_2^* = \begin{pmatrix} \tilde{\m J}_{2}^{11}+\m P_1 & \tilde{\m J}_{2}^{12} \\ \tilde{\m J}_{2}^{21} & \tilde{\m J}_{2}^{22} \end{pmatrix}$ with the superscripts denoting block partitions corresponding to $\ve y_{1}$ and $\ve y_2$, so that
\begin{align}
	\alpha(\ve y_2) &= \exp\{ \langle \tilde{\ve v}_2^{}, \ve y_2^{} \rangle + \ve y_2^{T} \widetilde{\m W} \ve y_2 \} \int_{\ve y_1}\exp\left\{ \left\langle \ve \eta_2^*,  \ve y_{1,2} \right\rangle +  \ve y_{1,2}^{T}  \m P_{2}^* \ve y_{1,2} \right\} \nonumber,
\end{align}

where the integral is (unnormalized) joint Gaussian on $(\ve y_1, \ve y_2)^T$ with $\ve \mu = -\frac{1}{2}\m P_2^{*-1}\ve \eta_2^*$  and $\m \Lambda=-2\m P_2^*$. The block marginalization properties of Gaussian distributions gives:
\begin{align}
	\alpha(\ve y_2) &= \exp\{ \langle \tilde{\ve v}_2^{}, \ve y_2^{} \rangle + \ve y_2^{T} \widetilde{\m W} \ve y_2 \}\exp\left\{ \left\langle \ve \eta_2,  \ve y_{2} \right\rangle +  \ve y_{2}^{T}  \m P_{2} \ve y_{2} \right\} \nonumber,
\end{align}
with
\begin{align}
	\ve \eta_2 = \tilde{\ve h}_{2}^2 - \tilde{\m J}_{2}^{21}(\tilde{\m J}_{2}^{11}+\m P_1)^{-1}(\tilde{\ve h}_{2}^1 + \ve \eta_1) \nonumber \\ 
	\m P_{2} = \tilde{\m J}_{2}^{22}-\tilde{\m J}_{2}^{21}(\tilde{\m J}_{2}^{11}+\m P_1)^{-1}\tilde{\m J}_{2}^{12} \nonumber
\end{align}
Thus, the message passing on the linear dynamical system ends up as updates on the natural parameters:
\begin{align}
	\alpha(\ve y_2) &= \exp\left\{ \langle \tilde{\ve v}_2^{}+ \ve \eta_2, \ve y_2^{} \rangle + \ve y_2^{T} \left( \widetilde{\m W}+\m P_{2}\right) \ve y_2 \right\} \nonumber,
\end{align}
which is analogous to the Kalman filter updates. Similar update equations can be derived for the backward pass and the marginal posteriors are given by the normalized product of the forward and backward passes. Since the resulting distributions are Gaussian, it is easy to compute the expected sufficient statistics required in the inference step described above for $q(u_{1:T})$. In practice, we will cycle between these two inference steps until convergence, after which the M-step is carried out.

\paragraph{Learning}
After repeating the inference step until convergence, we perform stochastic gradient updates by maximizing the ELBO (Equation \eqref{eq:ELBO}) with respect to all the model parameters. In particular, to optimize the first term:
\begin{align*}
    \E_{q}\bigg[\sum_{t=1}^T\log p(\ve x_t\mid\ve s_t^{(1)}, ..., \ve s_t^{(N)})\bigg]
\end{align*}
we sample $\ve s_t^{(1:N)} \sim q(\ve s_t^{(1:N)}),\forall t \in (1,\dots,T) \eqsp,$ and parameterize the mixing function with a decoder neural network $\ve f(\cdot; \ve \theta)$:
\begin{align}
    p(\ve x_t\mid \ve s_t) &= \mathcal{N}(\ve x_t ; \ve f(\ve s_t; \ve \theta), \m R) \eqsp.
\end{align}

\section{Details on experiments on simulated data} \label{sec:apdx:simudetails}
\paragraph{Simulated data} We simulated 100K long time-sequences from the \modelname and computed the mean absolute correlation coefficient (MCC) between the estimated latent components and ground true independent components. The switching linear dynamical system was simulated to have two latent hmm states, one that induced strong mean reverting behaviour upon the linear dynamical system, and another with oscillatory dynamics. The dimension of the linear dynamical system state-space was also set to 2 (1 + independent component). The HMM transition matrix was close to diagonal with 0.99 probability of staying in current state and 0.01 probability of transitioning to the other state, at each time step of the 100k long sequence. The code at [redacted for anonymity] provides the exact simulation details. To illustrate the dimensionality reduction capabilities we considered two settings where the observed data dimension $M$, was either 12 or 24 and the number of independent components, $N$ was 3 and 6, respectively. Therefore the model consist of $N$ independent processes of Equation \eqref{eq:sld}. Observations were created by the mixing function (Eq. \eqref{eq:mixing}) and additive Gaussian diagonal noise. We considered four levels of mixing of increasing complexity by randomly initialized MLPs of the following number of layers: 1 (linear ICA), 2, 3, and 5. 

\paragraph{Training details} All the experiments were run on ten different randomly simulated data sets to compute error bars. The model parameters, including the mixing function, were estimated using the inference and learning algorithm described above. All parameters were trained in ordered to increase the ELBO of the model; Adam with learning rate 1e-2 was used. The number of layers in the decoder networks was set equal to the number of mixing layers for both \modelname and IIA-HMM benchmark. The number of layers in the encoder \modelname was always one more than that for the decoder. We suspect this extra nonlinearity in the encoder helped training since VAEs have tendency to over-emphasize learning the likelihood term, which this may have alleviated. The number of hidden units was set at 128 and 64 for the decoder and encoder respectively. In order to avoid local minima, we started training from 20 different inital seeds and chose the model that reached the highest ELBO, or likelihood. The models were trained on University of Helsinki SLURM cluster until convergence, which in practice was approximately 12 hours on most settings. All training was done on CPUs only. Memory used for a single model to be trained was 15G RAM.

\paragraph{Further discussion of results} 
One possible reason for the relatively poor performance of IIA-HMM on the simulated data experiment (Figure \ref{fig:sim}) was suspected to be loss of information that resulted from the PCA preprocessing step. We explored this in additional experiments where there was no dimension reduction: \modelname still outperforms IIA-HHM also in this setting, although the latter's performance is now improved for small dimensionality (in 3-dimensions: MCC avg. 0.4 for 3 mixing layers), though remains clearly below \modelname (3-dimensions: MCC avg. 0.7). IIA-HMM performance for dimensions above 6, even without dimension reduction, was very poor (MCC < 0.3) suggesting its poor performance is not solely due to PCA, bur rather likely due to it lacking observation noise model (unlike all the other models we considered) and simpler model of latent dynamics (original iVAE also has no latent dynamics model but was here supplied with the ground-truth HMM latent state thus giving it a substantial advantage over IIA-HMM).  

\begin{figure}
    \centering
    \includegraphics[width=0.5\columnwidth]{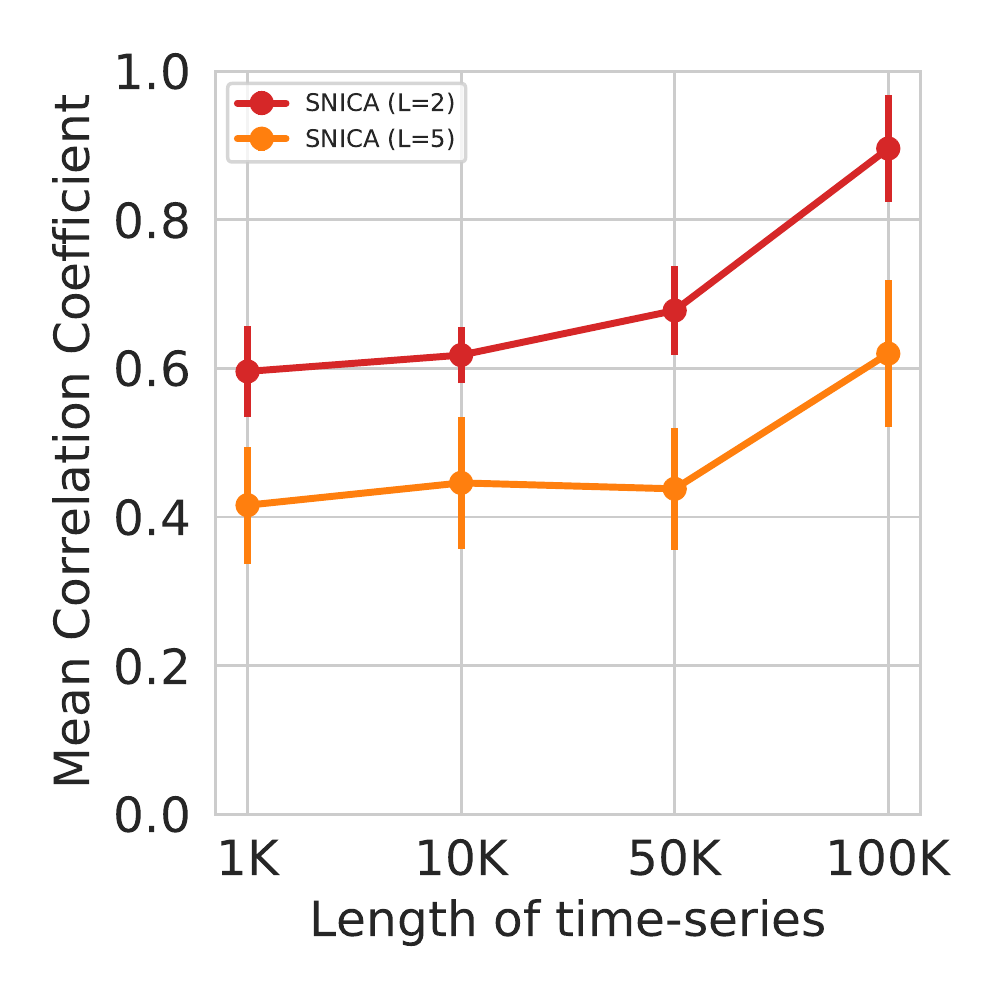}
    \caption{Mean absolute correlation coefficient between estimated and ground true independent components for varying lengths of training data for \modelname (N=3, M=12), for equal training time. Result shown for two different numbers of mixing layers L=2 and L=5}
    \label{fig:datasize_sim}
\end{figure}

\paragraph{Size of training data}
The theoretical identifiability results presented in this paper hold in the limit of infinite data. Hence, we hypothesized that the amount of training data may have large impact in any practical situations -- in addition to the usual benefits of increased dataset size. To explore this, we trained our model for varying lengths of datasets, with the results shown in Figure \ref{fig:datasize_sim}. We observed much better results for the largest dataset. Due to limited compute available to us, we leave it for future works to investigate even larger data sizes. 

\section{Details on MEG experiment} \label{detailmeg}
\paragraph{Data and Preprocessing}
The MEG data used were from the open Cam-CAN data repository\footnote{Acknowledgment for Cam-CAN data: Data collection and sharing for this project was provided by the Cambridge Centre for Ageing and Neuroscience (CamCAN). CamCAN funding was provided by the UK Biotechnology and Biological Sciences Research Council (grant number BB/H008217/1), together with support from the UK Medical Research Council and University of Cambridge, UK.} (available at http://www.mrc-cbu.cam.ac.uk/datasets/camcan/), and released under Creative Commons license. \citep{taylor2017cambridge, shafto2014cambridge}. The MEG dataset was collected using a 306-channel VectorView MEG system (Elekta Neuromag, Helsinki), consisting of 102 magnetometers and 204 orthogonal planar gradiometers with sampling 1000Hz. MEG data was preprocessed by temporal signal space separation (tsss; MaxFilter 2.2, Elekta Neuromag Oy, Helsinki, Finland) to remove noise from external sources and from HPI coils and head-motion was corrected (see \citep{taylor2017cambridge} for more details of the preprocessing). During the resting state recording, subjects sat still with their eyes closed for at least 8 min and 40 s. In the task-session data, the subjects carried out a (passive) audio–visual task including 120 trials of unimodal stimuli (60 visual stimuli: bilateral/full-field circular checkerboards; 60 auditory stimuli: binaural tones), presented at a rate of approximately 1 per second. In this study, We applied the method to 10 subjects' data and downsampled it to 128 Hz for saving computational resources, and only data from the planar gradiometers (204 channels) were used. We further band-pass filtered the data between 4 Hz and 30 Hz and normalized them to have zero-mean and unit variance. For the task-session data, we cropped each trial from -300ms to 600ms after the onset. The MNE package \citep{gramfort2013meg} was used for preprocessing.

\paragraph{SNICA setting}
We only used resting-state data for training. For saving memory, we selected 5-min long resting-state data from each subject. We temporally concatenated segments of each subject to form a dataset (5*60*128*10 = 384k time points) for training.  We fixed the number of independent components to 5, and set the number of hidden markov states and the dimension of the linear dynamical system to 2. The number of layesr in the encoder and decoder networks was set equal, and the number of hidden units was set to 32. Otherwise, all the settings were as in Simulation. 

\paragraph{Evaluation Methods}
For evaluation, we used the model trained with (unlabeled) resting-state data as feature extractors to perform a downstream task for classification of (labeled) task-session data. We carried out classification of the stimulus modality (auditory or visual) by using the estimated features. Classification was performed using a linear support vector machine (SVM) classifier trained on the stimulation modality labels and sliding-window-averaged features (width=10 and stride=3 samples) for each trial. The performance was evaluated by the generalizability of a classifier across subjects, i.e., one-subject-out cross-validation (OSO-CV). The hyperparameters of the SVM were determined by nested OSO-CV without using the test data. 
For comparison, IIA-HMM and IIA-TCL for the nonlinear vector autoregressive model (NVAM) were applied as baseline methods. Since IIA-HMM is not able to reduce the dimensionality, PCA was performed on the concatenated resting-state data to reduce the dimension to 5 for fair comparison. For IIA-TCL, we used segments of equal size, of length 10 s or 1280 data points, and also set the number of independent innovation to 5 for fair comparison.

We visualized the spatial patterns of the estimated features by plotting the weight vectors of units from encoder MLP in the topography map space. For the first layer, we have weight vectors (columns of the weight matrix $\ve W_1$) across sensors for each unit, and directly mapped them into brain topography space. And the weight matrix $\ve W_2$ multiplied by $\ve W_1$ to obtain weight vectors (columns of $\ve W_1 \ve W_2$) of sensors for each unit in the second layer, and so on for subsequent layers.

\paragraph{Interpreting the latent dynamics in the MEG experiments}
The learned parameters for the \modelname's latent dynamics, namely HMM-style switching, provide interesting interpretations in the MEG data experiment. Since we fix the number of HMM states to be two for each component, our assumption is that they can be interpreted as on/off or activity/inactivity. Such long-term on/off switching of the sources thus characterizes the nonstationary of the brain signal, as is quite often assumed in brain imaging. In particular, the components can be interpreted to represent different dynamic brain processes that are well-known to exist in the resting brain: visual, auditory, and other sensory networks; executive networks, attentional networks, and default mode network. The specific transition matrices for the hidden Markov discrete states can be interpreted to represent the movement between the transient brain states (process) in the real data. In particular, we found the HMM transition matrix to be close to diagonal, which suggests that we are capturing relatively slowly evolving states. The precise figures from the transition matrix suggest that on average a given state (active/inactive) lasts between 0.8 and 7 seconds. The marginal probabilities of the different states are fairly similar to each other, ranging between 0.3 to 0.6, thus all the states are relatively common in this sense. The hidden continuous states, on the other hand, are used here mainly as an algorithmic trick to easily model higher-order AR processes, and are thus harder to interpret.

\end{document}